\documentclass{article}

\usepackage{microtype}
\usepackage{graphicx}
\usepackage{subcaption}
\usepackage{booktabs} 

\usepackage{hyperref}
\usepackage{enumitem}

\newcommand{\MiB}{{\scriptsize MiB}}
\newcommand{\method}{NFM}

\usepackage[accepted]{icml2025}

\usepackage{amsmath}
\usepackage{amssymb}
\usepackage{mathtools}
\usepackage{amsthm}

\usepackage[capitalize,noabbrev]{cleveref}

\theoremstyle{plain}

\theoremstyle{definition}

\theoremstyle{remark}

\usepackage[textsize=tiny]{todonotes}

\icmltitlerunning{The Coupling Within: Flow Matching via Distilled Normalizing Flows}

\begin{document}

\twocolumn[
    \icmltitle{The Coupling Within: Flow Matching via Distilled Normalizing Flows}



    \icmlsetsymbol{equal}{*}

    \begin{icmlauthorlist}
        \icmlauthor{David Berthelot}{apple}
        \icmlauthor{Tianrong Chen}{apple}
        \icmlauthor{Jiatao Gu}{apple}
        \icmlauthor{Marco Cuturi}{apple}
        \icmlauthor{Laurent Dinh}{apple}
        \icmlauthor{Bhavik Chandna}{ucsd}
        \icmlauthor{Michal Klein}{apple}
        \icmlauthor{Josh Susskind}{apple}
        \icmlauthor{Shuangfei Zhai}{apple}
    \end{icmlauthorlist}

    \icmlaffiliation{apple}{Apple}
    \icmlaffiliation{ucsd}{UC San Diego}

    \icmlcorrespondingauthor{David Berthelot}{dberthelot@apple.com}

    \icmlkeywords{Machine Learning, ICML}

    \vskip 0.3in
]




\begin{abstract}
    Flow models have rapidly become the go-to method for training and deploying large-scale generators, owing their success to inference-time flexibility via adjustable integration steps.
    A crucial ingredient in flow training is the choice of \textit{coupling} measure for sampling noise/data pairs that define the \textit{flow matching} (FM) regression loss.
    While FM training defaults usually to independent coupling, recent works show that \textit{adaptive} couplings informed by noise/data distributions (e.g., via optimal transport, OT) improve both model training and inference.
    We radicalize this insight by shifting the paradigm: rather than computing adaptive couplings directly, we use \textit{distilled} couplings from a different, pretrained model capable of placing noise and data spaces in \textit{bijection}—a property intrinsic to \textit{normalizing flows} (NF) through their maximum likelihood and invertibility requirements.
    Leveraging recent advances in NF image generation via auto-regressive (AR) blocks, we propose Normalized Flow Matching (\method{}), a new method that distills the quasi-deterministic coupling of pretrained NF models to train student flow models.
    These students achieve the best of both worlds: significantly outperforming flow models trained with independent or even OT couplings, while also improving on the teacher AR-NF model.

\end{abstract}

\section{Introduction}
\label{introduction}
\begin{figure}[t]
    \centering
    \includegraphics[width=0.8\columnwidth, angle=-90, trim=0 16 0 16, clip]{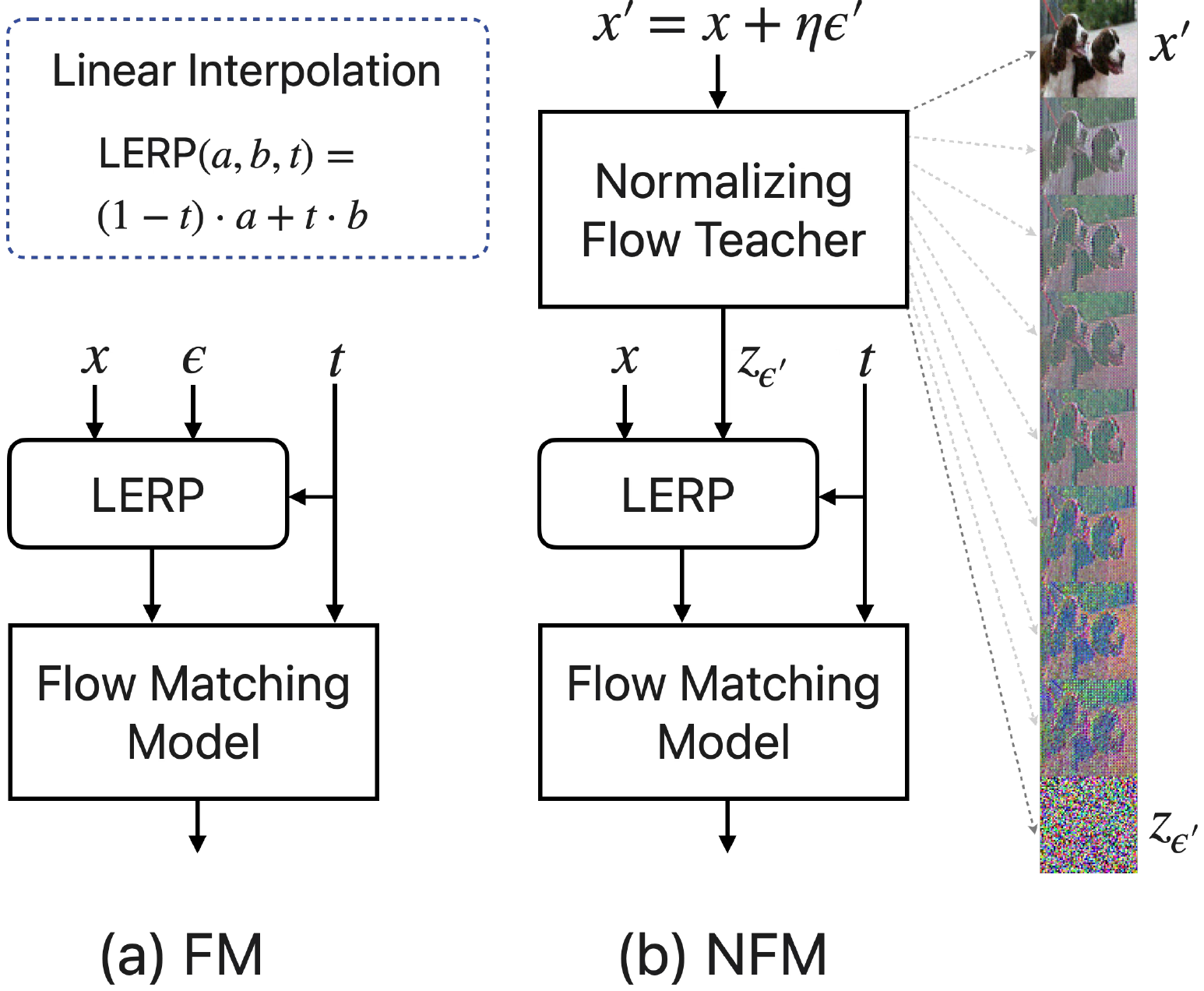}
    \caption{FM vs \method{} training}
    \label{pdf:method}
    \vskip -0.2in
\end{figure}

Flow Matching (FM) \cite{lipman2023flowmatching,albergo2023stochastic,peluchetti2022nondenoising,lipman2024flow} is a training paradigm proposed to learn Neural-ODE models~\citep{chen2018neural} so that they can flow from a distribution onto another, e.g. Gaussian to data. FM reflects the way flow models are used at inference time, in that the flow's time-parameterized velocity is trained to transport noise progressively towards data samples using a regression loss. At train time, the model is given pairs of noise and data samples; the velocity is trained to point from that noise towards data at any barycenter between the two. At inference time, that velocity is paired with an ODE integrator to recover outputs.

\textbf{On the coupling within flow matching.} A crucial element in this training procedure lies in the choice of a coupling measure used to sample noise/data pairs. While it is known that the independent coupling can, in theory, help the flow model recover a viable flow (one that satisfies the marginal requirement that flowing all possible noise vectors would recover all data points), this is rarely the case and that deficiency has encouraged iterative approaches such as rectified flows~\citep{liu2022flow}. Alternatively, several attempts have been made to substitute to the naive independent coupling a data informed one, using for instance ideas from optimal transport (OT)~\citep{tong2023improving,pooladian2023multisample,davtyan2025faster,zhang2025fitting,mousavi2025flow}. While these approaches can yield substantial improvements and better inference performance, they rely on the fairly simple intuition that flow models can be best trained when provided a \textit{good} coupling. This raises a very natural question: Can we explore a more sophisticated, data-informed and even \textit{better} approach to define the noise/data coupling used in FM? In other words, can the simple geometric considerations used so far in OT to define a coupling be superseded by an \textit{alternative} approach?

\textbf{Normalizing Flows.}
To obtain such an elusive coupling, we turn our attention to normalizing flows. Unlike flow models trained with flow matching, normalizing Flows (NF) \cite{tabak2010density,dinh2014nonlinear,rezende2015variational} do not suffer from any ambiguity when it comes to coupling noise to data: they learn, by construction, a \textit{bijection} between the data and a Gaussian noise source, and rely on the maximum likelihood principle to fit that bijection.
That bijectivity hinges on the fact that their architecture be completely \textit{invertible}.
In practice, their training is done by feeding data to a neural network that turns it into a noise vector, whose Gaussian negative log-likelihood (NLL) is minimized, thereby mapping data to noise.
At inference, the mapping from noise to data can be recovered by applying the network's inverse.
Recently, NF regained attention with Transformer Auto-Regressive Normalizing Flows, in short TarFlow \cite{zhai2025tarflow}, which achieved promising experimental results for generation and likelihood estimation \cite{gu2025starflow}.
The auto-regressive nature of these variants suffers from high latency sampling since each patch is generated sequentially for each block.


\textbf{NF for FM.} In this work, we propose a new method, Normalized Flow Matching (\method{}), that organically combines the best of both FM and NF worlds: we first train a NF model, which encodes a fast and bijective mapping from data to noise; a subsequent flow model is trained using FM based on the coupling encoded by the NF's map. We show that the resulting model manages to beat \textit{both} the FM and NF models trained from scratch.


More precisely, our contributions are:
\begin{itemize}[leftmargin=.2cm,itemsep=.0cm,topsep=0cm,parsep=2pt]
    \item \method{}, a simple coupling method that trains an FM model from the couplings produced by an NF teacher.
          The resulting students' sampling latency is orders of magnitude faster than the teacher's while, surprisingly, also significantly surpassing its FID.
    \item An analysis of structure of the NF Gaussian space, which, despite not having the same neighborhood properties as its input space, is facilitating FM convergence.
    \item In-depth experiments and ablations to tease apart the effects of couplings on convergence and FID.
\end{itemize}

\section{Preliminaries}

In the following, we denote an $n$-dimensional data sample as $x\in X\subseteq\mathbb{R}^n$ and its categorical label/class as $c\in\{0,...,k-1\}$ for a total of $k$ classes.
We denote a random Gaussian noise sample as $\epsilon\sim\mathcal{N}(0,I_n)$.
In the document we adopt a diffusion type of convention for notation consistency between NF and FM.
Consequently, $x$ refers to data, $\epsilon$ to noise.

\subsection{Flow Matching}
The training process of Flow Matching, as well as diffusion models, consists of first constructing a data distribution which interpolates random pairs of data and noise, which is known as the coupling process.
This is followed by a denoising objective that predicts the velocity term. Formally, for $t \in [0, 1]$, we have $x_t=(1-t)\cdot x+t\cdot\epsilon$, where $x \in X$, $\epsilon\sim\mathcal{N}(0,I_n)$.
Without loss of generality, in our diffusion convention $x_0=x$ and $x_1=\epsilon$.

The neural network $f_\text{FM}(x_t,c,t)$ is then trained with the loss:
\begin{equation} \label{eq:fm_loss}
    \mathcal{L}_\text{FM}=\Vert f_\text{FM}(x_t,c,t) - v_t\Vert_2^2\text{,}
\end{equation}
where $v_t=\epsilon-x$ is the velocity term.

Due to the stochastic nature of the construction of $x_t$, the optimal solution for Eq.\ref{eq:fm_loss} does not reach zero.
As a result, during inference, one needs to resort to an iterative procedure which starts from noise and gradually moves towards data. The inference procedure is defines as follows:
\begin{equation}
    x_1 \sim \mathcal{N}(0,I_n), \; x_{t-\Delta t} \;=\; x_t \;-\; \Delta t \, f_\text{FM}(x_t,c,t).
\end{equation}

\subsubsection{Noise/Data Coupling}
\label{sec:coupling}
One important design choice required to define the flow matching procedure lies in the choice of a noise/data coupling. Although the independent coupling (sampling independently Gaussian noise and data uniformly at random in the dataset) is the most standard \citep[\S4.8.3]{lipman2024flow}, one can benefit from selecting noise data pairs using a more careful pattern, with the intent of speeding training by assigning with a marginal-preserving coupling measure some noise regions to nearby data points.
That idea has been implemented to produce optimal transport-inspired couplings, starting with finite sample based approaches~\citep{tong2023improving,pooladian2023multisample,davtyan2025faster,zhang2025fitting} and culminating with Semi-Discrete Optimal-Transport (SD-FM) - the semidiscrete deterministic rule proposed by \citep{mousavi2025flow} to assign to any noise a specific point in the dataset using a lookup process. While these approaches can yield substantial improvements, they are model agnostic and can be best described as pre-processing steps that settle for specific rules to associate noise to data.

\subsection{Normalizing Flows}
The ideal coupling method should be one that directly maps a data point to a point from the Gaussian distribution, in which case the velocity prediction admits zero theoretical errors and sampling becomes one step. It turns out that there exists a family of learning algorithms whose objective serves this exact purpose: Normalizing Flows (NFs).

NFs are likelihood based methods trained with the change of variable formula. To be concrete, it learns an invertible function $f(x,c)$ which is optimized with the following loss:
\begin{equation} \label{eq:nf_loss}
    \mathcal{L}_\text{NF}=\frac{1}{2}\Vert f_{\text{NF}}(x,c) \Vert_2^2 - \log\left\vert\det \cfrac{\partial f_\text{\text{NF}}^{}}{\partial x}\right\vert\text{.}
\end{equation}

Historically, NFs are known to not scale well to large complex datasets.
Recently, this has changed with TarFlow~\cite{zhai2025tarflow} which used a Transformer based architecture for auto-regressive flows~\citep{papamakarios2017masked} (AF) to achieve competitive performance on image modeling tasks.
Beside the architectural improvements, another important ingredient of TarFlow is that it applies a small amount of Gaussian noise to the inputs, namely $x'=x+\eta\epsilon'$, where $\eta$ is a small value, before passing it to the network.
This is found to greatly boost the generalization of the trained model.

\subsection{Related works}

Accelerating sampling is a central research question for practical applications both for NF and FM.
In the NF context, while iterative approaches such as Jacobi iterations and speculative sampling~\citep{wiggers2020predictive, song2021accelerating, chen2023accelerating, monea2023pass, teng2025accelerating, liu2025accelerate} provide a training-free strategy to accelerate decoding, distillation approaches~\citep{hinton2014dark,lu2025bidirectional} yield a more consistent and significant speedup for normalizing flows~\citep{oord2018parallel, huang2020probability, baranchuk2021distilling, walton2025distilling}, as long as an additional computational time is spent on training the distilled model.

Other approaches focus on creating a hybrid model between NF and FM: Diffusion Normalizing Flow (DiffFlow) \cite{zhang2021diffusion} unifies normalizing flows and diffusion models by learning both the forward and backward stochastic differential equations jointly, minimizing the KL divergence between their trajectory distributions.
DiNof \cite{zand2024diffusion} introduces a hybrid generative model that combines stochastic diffusion processes with deterministic normalizing flows, using the flow network to parameterize a data-dependent prior at an intermediate diffusion timestep rather than relying solely on Gaussian noise.

Our method belongs to each category: it can be viewed both as a distillation of the NF model resulting in lower latency and better FID and as a hybrid model combining NF and FM in possibly the most straightforward manner.

\section{Method}

\subsection{Motivation}

As noted in Sec.\ref{sec:coupling}, OT finds an assignment between all data samples and all regions of the Gaussian space.
We build on the observation that NF is performing a similar task of mapping data samples to the Gaussian space.
In other words, our method is built on the natural synergy between distilling a NF model and learning effectively a diffusion model where the noise is essentially determined by the sample itself.

One particularly interesting point is that contrary to OT which, as the name says, optimally maps between the distributions, the learned NF mapping makes no claim about optimality and is likely sub-optimal since it is determined by the neural network capacity to model the data.
However, as will be shown in the experiments section, it turns out that such a sub-optimal mapping can outperform OT in the class-conditional setting.

\subsection{\method{}}

Our method distills a pretrained and frozen NF teacher $f_\text{NF}$ to a student $g$ which can be any architecture and, unlike the teacher, does not have to be invertible.

The training procedure mirrors the FM one, the difference being that instead of using random noise $\epsilon\sim\mathcal{N}(0,I_n)$, we use the Gaussian representation $z_{\epsilon'}\in\mathbb{R}^n$ produced by the teacher model:
$$z_{\epsilon'}=f_\text{NF}(x+\eta\epsilon',c)/\sigma_f\text{,}$$
where $\epsilon'\sim\mathcal{N}(0,I_n)$, $\eta$ is the regularization noise amount originally used when training $f_\text{NF}$ and $\sigma_f\in\mathbb{R}^n$ is the teacher's output normalization vector that enforces expected unit norm, e.g., $\sigma_f^2=\mathbb{E}_{(x,c)\sim X,\epsilon'\sim\mathcal{N}(0,I_n)}[f_\text{NF}(x+\eta\epsilon',c)^2]$.

As previously mentioned, we simply substitute $z_{\epsilon'}$ to $\epsilon$, consequently the loss for FM distillation is:
$$\mathcal{L}_\text{FM}=\Vert g((1-t)\cdot x+t\cdot z_{\epsilon'},c,t) - (z_{\epsilon'} - x)\Vert_2^2\text{.}$$

This substitution does \emph{not} change the Flow Matching objective: for a fixed noise schedule, FM remains the same weighted-ELBO/path-KL regression of a vector field along the linear interpolant.
We only replace the endpoint coupling $(x,\epsilon)$ with $(x,z_{\epsilon'})$ where $z_\epsilon$ is normalized to be approximately $\mathcal{N}(0,I_n)$ marginally.
Accordingly, the training target is still the endpoint difference $v_t=x_\text{source}-x_\text{target}$, here $v_t=z_{\epsilon'}-x$, while the time-dependent weighting is unchanged because it depends solely on the chosen schedule.

Comparing the maximum training noise added in our method with the one added in FM, we find that $t=\eta/(1+\eta)$ by converting from the variance exploding setting to the mean preserving one.
Since in practice TarFlow uses a small value for $\eta$, our method effectively trains Flow Matching with much less noise.
To illustrate this with an example, for ImageNet64 TarFlow uses $\eta=0.05$ resulting in maximum flow matching noise level of $0.0476$ in our method while the original flow matching maximum noise is $1.0000$, potentially making the inference path straighter.
Beyond using a lower maximum noise level (as quantified above), the teacher-induced coupling typically reduces the \emph{conditional velocity variance} $\mathrm{Var}(v_t\mid x_t,t)$ which in our case is $v_t=z_{\epsilon'}-x$ as to opposed to $v_t=\epsilon-x$ in vanilla FM.
This yields (i) lower target/gradient variance and thus more stable sample-efficient optimization, and (ii) a smoother conditional-mean field $u_q(x_t,t)=\mathbb{E}[v_t\mid x_t,t]$, which empirically reduces trajectory curvature and improves ODE integration stability, often requiring fewer steps.

As is typically done for both NF and FM to make guidance work, we randomly drop the class label by setting it to $\varnothing$ with probability $p$ during training.

Generative sampling for \method{} is the same as for FM.


\subsection{Normalizing Flows z-space structure}
Normalizing Flows project an input $x'\in\mathbb{R}^n$ to a Gaussian representation $z_{\epsilon'}\in\mathbb{R}^n$.
One question that arises is where are the points projected exactly: do neighbors remain neighbors after projection?
Surprisingly the answer is no, at least in the case of TarFlow.
We cannot comment about other NF designs as studying all of them is beyond the scope of this paper.

Let's recall that $x'=x+\eta\epsilon'$.
We use the distance $d(a,b)=\Vert a - b \Vert_2/\sqrt{2n}$ to define:
\begin{align*}
    d_x(x_1,\epsilon_1,x_2,\epsilon_2) & =d(x_1 + \eta \epsilon_1, x_2 + \eta \epsilon_2)\text{,}               \\
    d_z(x_1,\epsilon_1,x_2,\epsilon_2) & =d(f(x_1 + \eta \epsilon_1,c_1), f(x_2 + \eta \epsilon_2,c_2))\text{.}
\end{align*}

Tab.\ref{table:z-space},\ref{table:z-space256} present the average distances in input $x$-space and output $z$-space for various scenarii on the entire dataset for ImageNet64 and ImageNet256.

The first group $x_1\neq x_2, \epsilon_1\neq\epsilon_2$, represents the average distances between different samples and their representations with different noises.
We observe that $d_z=1$ in this case, since the NF learns $z$ representations that are on the surface of a Gaussian unit sphere.
The $x$ representations get further apart as expected when more noise is added, e.g. $d_x$ increases with $\eta$.

The second group $x_1=x_2, \epsilon_1\neq\epsilon_2$, represents how spread apart are the representations for each image.
We see that $d_x=\eta$ which confirms representations of $x$ are at distance $\eta$ of $x$ as expected.
More interestingly, the $d_z$ columns shows that the $z$ representations are much more spread apart on the unit sphere than they are initially in the $x$-space.
For example, in Tab.\ref{table:z-space} for $\eta=0.05$, the corresponding distance between representations is $0.85$ which is pretty high considering $1.00$ is the perfect Gaussian sphere coverage.
Put another way, a single sample's representations cover a large area of the Gaussian sphere.
And as expected, the larger $\eta$, the larger $d_z$.

Finally in group 3, we look at the distances between different images under the same noise, e.g. $x_1\neq x_2, \epsilon_1=\epsilon_2$.
The $d_x$ column in this case simply captures the distance between images: it is fixed and only depends on the dataset and the normalization of the input space.
More interesting is the $d_z$ column which measures the distance between image representations.
First we observe that the greater $\eta$, e.g. the more noise we add, the lower $d_z$, e.g. the closer images representations are squeezed together.
Second, for the values of $\eta$ which yield best FIDs, as will be seen in the next section, we observe that $d_x=d_z$.
This is unexpected, and we do not know whether it is pure coincidence or not.
Third, we finally observe that $d_z(x_1=x_2, \epsilon_1\neq\epsilon_2) > d_z(x_1\neq x_2, \epsilon_1=\epsilon_2)$, in other words image representations are closer to other images representations under the same noise than they are to themselves under different noises.
This is the most counter-intuitive observation of all, as the many representations of the same image are not its nearest neighbors in $z$-space while they are in $x$-space.

As we will see experimentally in the next section, despite the unexpected structure of the NF's $z$-space, using a normalizing flow coupling still yields large gains in convergence and few-step FID for Flow Matching.

\begin{table}[t]  
    \caption{$z$-space structure for ImageNet64}
    \label{table:z-space}
    \vskip 0.15in
    \begin{center}
        \begin{small}
            \begin{sc}
                \begin{tabular}{lcccccc}
                    \toprule
                            & \multicolumn{2}{c}{$x_1\neq x_2$}              & \multicolumn{2}{c}{$x_1=x_2$}                  & \multicolumn{2}{c}{$x_1\neq x_2$}                                      \\
                            & \multicolumn{2}{c}{$\epsilon_1\neq\epsilon_2$} & \multicolumn{2}{c}{$\epsilon_1\neq\epsilon_2$} & \multicolumn{2}{c}{$\epsilon_1=\epsilon_2$}                            \\
                    $\eta$  & $d_x$                                          & $d_z$                                          & $d_x$                                       & $d_z$  & $d_x$  & $d_z$  \\
                    \midrule
                    $0.025$ & $0.50$                                         & $1.00$                                         & $0.025$                                     & $0.78$ & $0.50$ & $0.67$ \\
                    $0.05$  & $0.50$                                         & $1.00$                                         & $0.05$                                      & $0.85$ & $0.50$ & $0.57$ \\
                    $0.1$   & $0.51$                                         & $1.00$                                         & $0.10$                                      & $0.90$ & $0.50$ & $0.47$ \\
                    $0.2$   & $0.54$                                         & $1.00$                                         & $0.20$                                      & $0.94$ & $0.50$ & $0.37$ \\
                    \bottomrule
                \end{tabular}
            \end{sc}
        \end{small}
    \end{center}
    \vskip -0.1in
\end{table}


\begin{table}[t]  
    \caption{$z$-space structure for ImageNet256}
    \label{table:z-space256}
    \vskip 0.15in
    \begin{center}
        \begin{small}
            \begin{sc}
                \begin{tabular}{lcccccc}
                    \toprule
                            & \multicolumn{2}{c}{$x_1\neq x_2$}              & \multicolumn{2}{c}{$x_1=x_2$}                  & \multicolumn{2}{c}{$x_1\neq x_2$}                                      \\
                            & \multicolumn{2}{c}{$\epsilon_1\neq\epsilon_2$} & \multicolumn{2}{c}{$\epsilon_1\neq\epsilon_2$} & \multicolumn{2}{c}{$\epsilon_1=\epsilon_2$}                            \\
                    $\eta$  & $d_x$                                          & $d_z$                                          & $d_x$                                       & $d_z$  & $d_x$  & $d_z$  \\
                    \midrule
                    $0.025$ & $0.80$                                         & $1.00$                                         & $0.025$                                     & $0.26$ & $0.80$ & $0.99$ \\
                    $0.050$ & $0.80$                                         & $1.00$                                         & $0.05$                                      & $0.37$ & $0.80$ & $0.97$ \\
                    $0.100$ & $0.80$                                         & $1.00$                                         & $0.10$                                      & $0.50$ & $0.80$ & $0.93$ \\
                    $0.250$ & $0.83$                                         & $1.00$                                         & $0.25$                                      & $0.69$ & $0.80$ & $0.80$ \\
                    $0.300$ & $0.85$                                         & $1.00$                                         & $0.30$                                      & $0.72$ & $0.80$ & $0.76$ \\
                    $0.400$ & $0.89$                                         & $1.00$                                         & $0.40$                                      & $0.77$ & $0.80$ & $0.70$ \\
                    $0.500$ & $0.94$                                         & $1.00$                                         & $0.50$                                      & $0.81$ & $0.80$ & $0.64$ \\
                    \bottomrule
                \end{tabular}
            \end{sc}
        \end{small}
    \end{center}
    \vskip -0.1in
\end{table}


\section{Experiments}
\label{sec:experiments}

We used ImageNet at resolutions 64 and 256 for our experiments.
Resizing was done using the $\texttt{PIL}$ library box resize method as in \cite{chrabaszcz2017downsampled}.
For ImageNet256, we use a pretrained variational auto-encoder from Stable Diffusion \cite{rombach2022highresolution}\footnote{\url{https://huggingface.co/stabilityai/sd-vae-ft-mse}} to transform the data from pixel space to latent space.
The probability to drop a class label is set to $p=0.1$ for all methods, including teacher training.
Pixel images are represented on the $[-1,1]$ interval.
For consistency, we trained TarFlow teachers, students and baselines using our own re-implementations and on the same datasets \footnote{\url{https://github.com/apple/ml-nfm}}.
All of our experiments are in the class-conditional setting.

The TarFlow teacher models are named as TF-$blocks\times depth$-$channels/patch$, for example TF-6$\times$2+2$\times$26-768/2 is a TarFlow model with $6$ meta-blocks of depth $2$ and $2$ meta-blocks of depth $26$ and patch size $2$.
Teachers are trained for $512$\MiB{} (Mebibytes, e.g. $1{\scriptstyle\text{MiB}}=2^{20}=1\,048\,576$) samples ($\sim420$ epochs).

All FM students and baseline runs are trained for $256$\MiB{} samples ($\sim210$ epochs).
Noise levels $t$ are sampled from the logit-normal distribution \cite{aitchison1982statistical,esser2024scaling}  $\text{lognorm}(a,b)=\text{sigmoid}(a + b\cdot\mathcal{N}(0,1))$ with $a=-0.2$ and $b=1$.
Sampling uses either Euler \cite{euler1768institutionum,song2021denoising} or Heun \cite{heun1900neue,karras2022elucidating} solvers: Euler when $\text{NFE}\leq 5$ and Heun when $\text{NFE}\geq 5$ (since $\text{NFE}=2\times\text{steps}-1$), where NFE denotes the number of function evaluations.
Unless stated otherwise, we use the schedule $t^2=\{1^2,(1-\delta t)^2,\ldots,(\delta t)^2\}$ with $\delta t=1/\text{steps}$.

Following commonly accepted practice, all reported Fréchet Inception Distance (FID) \cite{dowson1982frechet} results are measured between the training dataset distribution and $50\,000$ randomly generated samples.
Unless stated otherwise all reported FIDs numbers use guidance.
The actual guidance value is determined per experiment (model, sampler, NFEs) by means of a Golden Section Search \cite{kiefer1953sequential}.
For computational efficiency we do a $12$ iterations search using a quick FID approximation by computing it on only $8\,192$ randomly generated samples.
The optimal guidance parameter is ultimately rounded to two significant digits and is then used to generate the $50\,000$ samples required to compute the reported FID.

\subsection{SotA comparisons}
Comparisons can be found in Tab.\ref{table:fid-64},\ref{table:fid-256} where FID at various training points are shown.
For easier and more detailed visualization, the FIDs are also presented as curves as function of training samples seen in Fig.\ref{plot:conv-fm64},\ref{plot:conv-fmf256}.

From a distillation perspective, the student can match and even outperform the FID of the teacher, while being faster and using fewer weights.
The student outperforming the teacher is, however, surprising.

\subsubsection{Convergence analysis and couplings}
\begin{figure}[ht]
    \vskip 0.2in
    \begin{center}
        \includegraphics[width=\columnwidth]{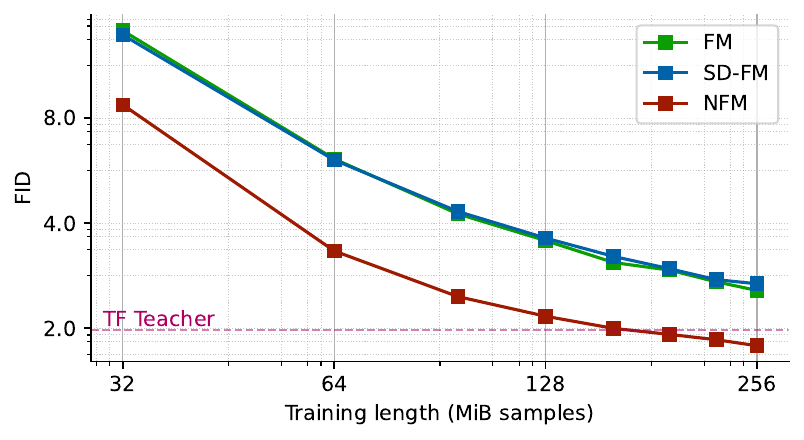}
        \caption{FID convergence on ImageNet64}
        \label{plot:conv-fm64}
    \end{center}
    \vskip -0.2in
\end{figure}

\begin{figure}[ht]
    \vskip 0.2in
    \begin{center}
        \includegraphics[width=\columnwidth]{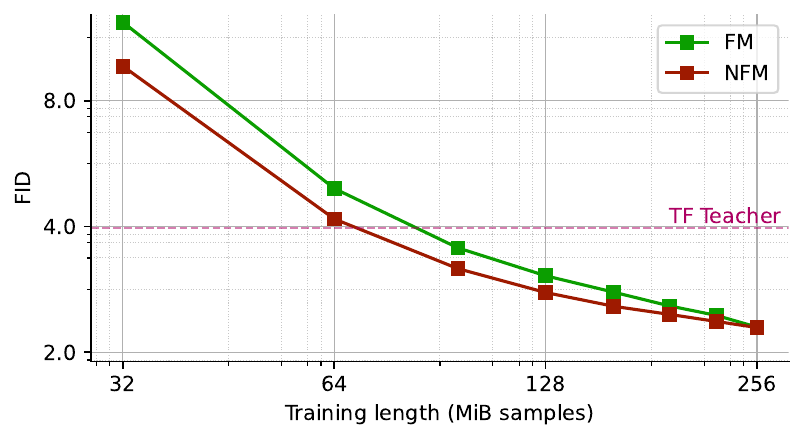}
        \caption{FID convergence on ImageNet256 (31 NFE sampling)}
        \label{plot:conv-fmf256}
    \end{center}
    \vskip -0.2in
\end{figure}

We also verify experimentally that data-noise coupling should accelerate the training of diffusion models in Tab.\ref{table:fid-64},\ref{table:fid-256} and Fig.\ref{plot:conv-fm64},\ref{plot:conv-fmf256}.
Both SD-FM and TarFlow couplings accelerate convergence resulting in better FID values earlier on than the random noise baseline.
This effect is particularly noticeable with less NFEs, like $15$ and $7$, where both methods achieve significant gains over the vanilla FM.
TarFlow couplings performs particularly well, yielding FIDs that outperforms both by a large margin.
We want to make it clear that we are not opposing \method{} and SD-FM, we see them as potentially complementary techniques.
A deeper discussion on the couplings differences and complementarity can be found in Sec.\ref{sec:label-ot}.

In Tab.\ref{table:kappa-64}, we report the curvature \cite{lee23minimizing} computed on $50\,000$ randomly generated samples without CFG.
Euler($t$) indicates applying the Euler solver to a linear schedule $t=\{1,1-\delta t,\ldots,\delta t\}$ with $\delta t=1/\text{steps}$ and Heun($t^2$) indicates applying the Heun solver to the square schedule $t^2$ already mentioned earlier.
$x_1,\ldots,x_m$ with $x_i\in\mathbb{R}^n$ are the samples produced by the solver and $v^{(i)}_t$ is the velocity predicted by the FM neural network to denoise from $x_i$ to $x_{i+1}$.
The curvature is computed as follows:
$\kappa=\mathbb{E}_i[\Vert x_1 - x_m - v^{(i)}_t\Vert^2_2]/n$.
By definition the curvature is not comparable across solvers and timestep schedules.
We observe for each solver/schedule that \method{} produces significantly straighter paths than FM and SD-FM.

\begin{table}[t]
    \caption{Curvature $\kappa$ on ImageNet64}
    \label{table:kappa-64}
    \vskip 0.15in
    \begin{center}
        \begin{small}
            \begin{sc}
                \begin{tabular}{lcccc}
                    \toprule
                    Solver      & NFE & FM     & SD-FM  & \method{}       \\
                    \midrule
                    Heun($t^2$) & 31  & 0.0864 & 0.0767 & \textbf{0.0435} \\
                    Euler($t$)  & 128 & 0.0386 & 0.0289 & \textbf{0.0181} \\
                    \bottomrule
                \end{tabular}
            \end{sc}
        \end{small}
    \end{center}
    \vskip -0.1in
\end{table}

For ImageNet256, we only report the Gaussian noise baseline and TarFlow couplings.
Like for ImageNet64, TarFlow couplings accelerate convergence resulting in lower FIDs and gains are particularly significant with less NFEs.


\begin{table}[t]
    \caption{FID comparisons ImageNet64 with SiT-XL/4 for TarFlow teacher TF-6$\times$2+2$\times$26-1024/2 with FID=$1.98$.} 
    \label{table:fid-64}
    \vskip 0.15in
    \begin{center}
        \begin{small}
            \begin{sc}
                \begin{tabular}{lcr@{\hspace{8pt}}r@{\hspace{8pt}}r@{\hspace{8pt}}r@{\hspace{8pt}}}
                    \toprule
                    Method    & NFE  & \multicolumn{4}{c}{FID}                                                       \\
                              &      & $32$\MiB{}              & $64$\MiB{}      & $128$\MiB{}     & $256$\MiB{}     \\
                    \midrule
                    FM        & $31$ & $14.33$                 & $6.11$          & $3.56$          & $2.57$          \\ 
                    SD-FM     & $31$ & $13.83$                 & $6.06$          & $3.63$          & $2.68$          \\ 
                    \method{} & $31$ & $\textbf{8.73}$         & $\textbf{3.33}$ & $\textbf{2.17}$ & $\textbf{1.78}$ \\ 
                    \midrule
                    FM        & $15$ & $17.15$                 & $7.47$          & $5.24$          & $4.80$          \\ 
                    SD-FM     & $15$ & $14.45$                 & $6.39$          & $3.98$          & $3.15$          \\ 
                    \method{} & $15$ & $\textbf{9.47}$         & $\textbf{3.65}$ & $\textbf{2.53}$ & $\textbf{2.15}$ \\ 
                    \midrule
                    FM        & $7$  & $20.33$                 & $14.87$         & $12.74$         & $13.01$         \\ 
                    SD-FM     & $7$  & $17.02$                 & $9.61$          & $7.10$          & $6.41$          \\ 
                    \method{} & $7$  & $\textbf{11.01}$        & $\textbf{5.03}$ & $\textbf{3.76}$ & $\textbf{3.23}$ \\ 
                    \bottomrule
                \end{tabular}
            \end{sc}
        \end{small}
    \end{center}
    \vskip -0.1in
\end{table}

\begin{table}[t]
    \caption{FID vs NFEs comparisons for ImageNet64 with SiT-XL/4 for TarFlow teacher TF-6$\times$2+2$\times$26-1024/2 with FID=$1.98$.} 
    \label{table:nfe-64}
    \vskip 0.15in
    \begin{center}
        \begin{small}
            \begin{sc}
                \begin{tabular}{lcrrr}
                    \toprule
                    Solver & NFE & FM    & SD-FM & \method{}     \\ 
                    \midrule
                    Heun   & 31  & 2.57  & 2.68  & \textbf{1.78} \\
                    Heun   & 15  & 4.80  & 3.15  & \textbf{2.15} \\
                    Heun   & 7   & 13.01 & 6.41  & \textbf{3.23} \\
                    Heun   & 5   & 17.56 & 9.29  & \textbf{4.01} \\
                    Euler  & 5   & 21.05 & 12.18 & \textbf{3.92} \\
                    Euler  & 4   & 21.81 & 13.76 & \textbf{4.47} \\
                    Euler  & 3   & 40.47 & 20.69 & \textbf{6.34} \\
                    \bottomrule
                \end{tabular}
            \end{sc}
        \end{small}
    \end{center}
    \vskip -0.1in
\end{table}

\begin{table}[t]
    \caption{FID comparisons ImageNet256 with SiT-XL/2 for TarFlow teacher TF-6$\times$2+2$\times$26-1024/1 with FID=$3.96$.} 
    \label{table:fid-256}
    \vskip 0.15in
    \begin{center}
        \begin{small}
            \begin{sc}
                \begin{tabular}{lcrrrr}
                    \toprule
                    Method    & NFE  & \multicolumn{4}{c}{FID}                                                       \\
                              &      & $32$\MiB{}              & $64$\MiB{}      & $128$\MiB{}     & $256$\MiB{}     \\
                    \midrule
                    FM        & $31$ & $12.33$                 & $4.93$          & $3.05$          & $2.30$          \\ 
                    \method{} & $31$ & $\textbf{9.68}$         & $\textbf{4.17}$ & $\textbf{2.78}$ & $\textbf{2.29}$ \\ 
                    \midrule
                    FM        & $15$ & $12.40$                 & $5.07$          & $3.37$          & $2.84$          \\ 
                    \method{} & $15$ & $\textbf{10.36}$        & $\textbf{4.14}$ & $\textbf{2.84}$ & $\textbf{2.45}$ \\ 
                    \midrule
                    FM        & $7$  & $19.92$                 & $13.72$         & $13.60$         & $12.41$         \\ 
                    \method{} & $7$  & $\textbf{10.99}$        & $\textbf{5.14}$ & $\textbf{3.93}$ & $\textbf{3.43}$ \\ 
                    \bottomrule
                \end{tabular}
            \end{sc}
        \end{small}
    \end{center}
    \vskip -0.1in
\end{table}

\begin{table}[t]
    \caption{FID vs NFEs comparisons for ImageNet256 with SiT-XL/2 for TarFlow teacher TF-6$\times$2+2$\times$26-1024/1 with FID=$3.96$.} 
    \label{table:nfe-256}
    \vskip 0.15in
    \begin{center}
        \begin{small}
            \begin{sc}
                \begin{tabular}{lcrrrr}
                    \toprule
                    Solver & NFE & \multicolumn{2}{c}{FM} & \multicolumn{2}{c}{\method{}}                                 \\ 

                           &     & $128$\MiB{}            & $256$\MiB{}                   & $128$\MiB{}   & $256$\MiB{}   \\
                    \midrule
                    Heun   & 31  & 3.05                   & 2.30                          & \textbf{2.78} & \textbf{2.29} \\
                    Heun   & 15  & 3.37                   & 2.84                          & \textbf{2.84} & \textbf{2.45} \\
                    Heun   & 7   & 13.60                  & 12.41                         & \textbf{3.93} & \textbf{3.43} \\
                    Heun   & 5   & 48.51                  & 48.73                         & \textbf{6.43} & \textbf{5.75} \\
                    Euler  & 5   & 11.60                  & 10.48                         & \textbf{4.72} & \textbf{4.04} \\
                    Euler  & 4   & 37.23                  & 33.63                         & \textbf{6.94} & \textbf{6.10} \\
                    \bottomrule
                \end{tabular}
            \end{sc}
        \end{small}
    \end{center}
    \vskip -0.1in
\end{table}

\subsubsection{Latency gains}

As shown in Tab.\ref{table:latency-64}, the distilled networks have lower latency than the original teacher, as is expected.
Surprisingly, not only can the student outperform the teacher's FID as previously mentioned but it also achieves a 32$\times$ faster latency.

\begin{table}[t]
    \caption{Latency vs FID ImageNet64 with SiT-XL/4 students and TF-6$\times$2+2$\times$26-1024/2 teacher.}
    \label{table:latency-64}
    \vskip 0.15in
    \begin{center}
        \begin{small}
            \begin{sc}
                \begin{tabular}{lrrr}
                    \toprule
                    Method  & FID    & Latency   & Speedup     \\
                    \midrule
                    TarFlow & $1.98$ & $10.7965$ & $1\times$   \\
                    NFM(31) & $1.78$ & $0.3376$  & $32\times$  \\
                    NFM(15) & $2.15$ & $0.1585$  & $68\times$  \\
                    NFM(7)  & $3.23$ & $0.0744$  & $145\times$ \\
                    \bottomrule
                \end{tabular}
            \end{sc}
        \end{small}
    \end{center}
    \vskip -0.1in
\end{table}

\subsubsection{Pairwise vs distributional distillation}
\begin{figure}[ht]
    \vskip 0.2in
    \centering
    \begin{subfigure}{1.\columnwidth}
        \includegraphics[width=\textwidth]{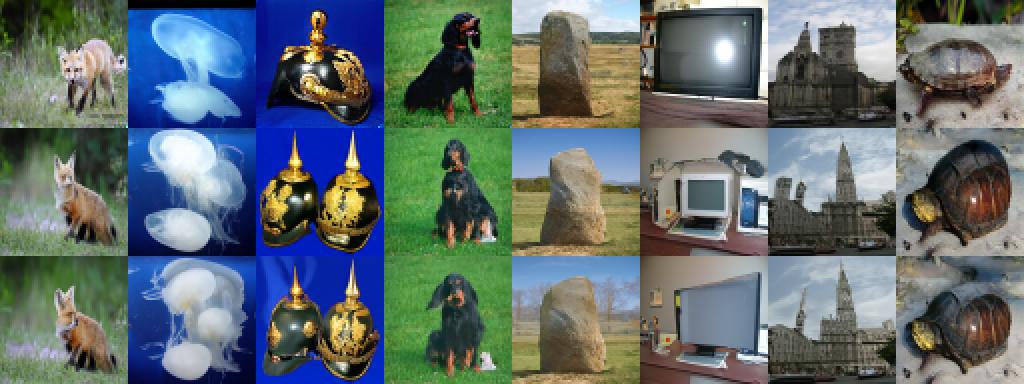}
        \caption{With CFG}
    \end{subfigure}
    \\
    \begin{subfigure}{1.\columnwidth}
        \includegraphics[width=\textwidth]{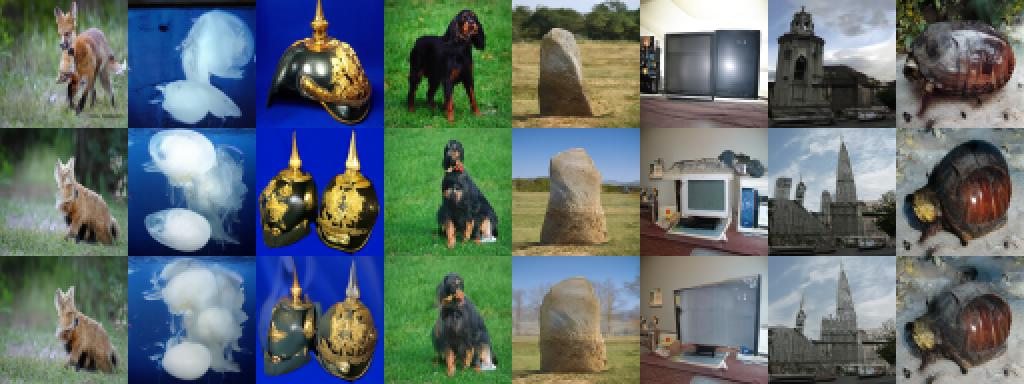}
        \caption{Without CFG}
    \end{subfigure}
    \caption{ImageNet64 same-seed random samples. Rows 1,2,3 are resp. from TF teacher, FM(31) student and FM(7) student.}
    \label{image:same-seed64}
    \vskip -0.2in
\end{figure}

Fig.\ref{image:same-seed64} shows random samples using the same seeds to sample from the ImageNet64 models.
Rows $1,2,3$ contain resp. samples from the TF teacher, and the FM(31) and FM(7) students with respective FIDs ($1.98,1.78,3.23$) from Tab.\ref{table:fid-64}.
Labels were manually picked to present diverse cases.

While the loss we used in our method should result in pair-wise matching, it is clear from the samples that it is only loosely the case.
Indeed, the samples while often sharing some similarity are still different from each other.
The outcome is that in practice the distillation behaves as distributional matching.

While we cannot say for sure what causes this phenomenon, multiple conjectures can be considered:
\begin{itemize}[leftmargin=.2cm,itemsep=.0cm,topsep=0cm,parsep=2pt]
    \item The teacher and student inductive biases are different due to the choice of different architectures.
    \item The FM sampling itself does introduce noise.
\end{itemize}

\subsection{Teacher impact}
\label{sec:teacher-impact}
We trained various teachers by changing the patch size and the network's number of channels.
For each teacher, we trained a corresponding distilled student for FM.
In Tab.\ref{table:teacher-64}, we report the model size, negative likelihood (NLL) and FIDs for the teacher and student.
It can be observed that the FID of the student correlates relatively well with the NLL of the teacher.
The teacher size or choice of sub-blocks seems less directly correlated to FID, although it obviously has an impact on the NLL.
Overfitting could explain why for the largest teacher, the correlation starts to break down.

The general takeaway is that better teachers from the perspective of NLL learn better Gaussian representations which in turn yield better performing students.
In particular, several of the teachers have the same number of parameters but their respective NLL correlates directly with the FID obtained post-distillation.

\begin{table}[t]
    \caption{TarFlow teacher impact on FM student for ImageNet64, T/S columns for FID contain Teacher and Student FIDs}
    \label{table:teacher-64}
    \vskip 0.15in
    \begin{center}
        \begin{small}
            \begin{sc}
                \begin{tabular}{lrrr}
                    \toprule
                    Teacher                          & Size    & NLL     & FID (T/S)     \\
                    \midrule
                    TF-6$\times$2+2$\times$26-768/4  & $460$K  & -2.1252 & $3.93 / 2.67$ \\ 
                    TF-6$\times$2+2$\times$26-768/2  & $460$K  & -2.1410 & $2.14 / 1.80$ \\ 
                    TF-6$\times$2+2$\times$26-1024/2 & $823$K  & -2.1500 & $1.98 / 1.78$ \\ 
                    TF-6$\times$2+2$\times$26-1536/2 & $1839$K & -2.1691 & $2.21 / 1.84$ \\ 
                    \midrule
                    TF-8$\times$8-768/2              & $460$K  & -2.1353 & $3.42 / 2.41$ \\ 
                    \bottomrule
                \end{tabular}
            \end{sc}
        \end{small}
    \end{center}
    \vskip -0.1in
\end{table}

\subsubsection{Noise impact}

Let us recall that TarFlow models used as teachers are trained with a certain amount of noise $\eta$ added to the input $x$ to prevent infinite NLL and NaNs during the training process.
When doing image generation for TarFlow, the noisified image $x'$ is generated from $x+\eta\epsilon'=x'=f^{-1}(z,c)$.
The noise $\epsilon'$ must be removed from $x'$ to obtain $x$.
This is done classically in TarFlow by a gradient descent step, $\epsilon=\nabla_{x'} f(x',c)$.
Consequently TarFlow training involves finding a good $\eta$ that yields low FID samples.

However, in the context of distillation we use only TarFlow's forward pass $f$ but not $f^{-1}$ nor its denoising process.
Therefore the rules for picking $\eta$ are less clear: how do we pick a good $\eta$ that yields low FID samples once the teacher is distilled?
In practice, we picked $\eta$ following TarFlow's paper to train teachers optimized towards sampling.

In this section, we take a closer look at what happens when we train the same TarFlow model with various $\eta$ and distill them.
Tab.\ref{table:noise-64} shows the results.
Interestingly, and conveniently, it appears that the best $\eta$ for teacher sampling are also the best ones for teacher distillation.
This seems to point to a much more fundamental role for $\eta$ than just preventing infinite NLL, as it connects to the generation quality of both the teacher and of the student.

It is also worth noting that, as expected, the final NLL is greatly influenced by $\eta$.
However when varying $\eta$ FID no longer correlates with NLL.
Consequently the correlation observed between NLL and FID in Sec.\ref{sec:teacher-impact} appears restricted to the same fixed values of $\eta$.



\begin{table}[t]
    \caption{TF-6$\times$2+2$\times$26-1024/2 teacher noise impact on FM student for ImageNet64, T/S columns for FID contain Teacher and Student FIDs.}
    \label{table:noise-64}
    \vskip 0.15in
    \begin{center}
        \begin{small}
            \begin{sc}
                \begin{tabular}{rrr}
                    \toprule
                    $\eta$  & NLL     & FID (T/S)     \\
                    \midrule
                    $0.025$ & -2.6223 & $2.40 / 1.85$ \\ 
                    $0.05$  & -2.1500 & $1.98 / 1.78$ \\ 
                    $0.1$   & -1.6040 & $1.95 / 1.84$ \\ 
                    $0.2$   & -1.0040 & $3.67 / 2.31$ \\ 
                    \bottomrule
                \end{tabular}
            \end{sc}
        \end{small}
    \end{center}
    \vskip -0.1in
\end{table}

\subsection{Unguided behavior}
\begin{figure}[ht]
    \vskip 0.2in
    \begin{center}
        \includegraphics[width=\columnwidth]{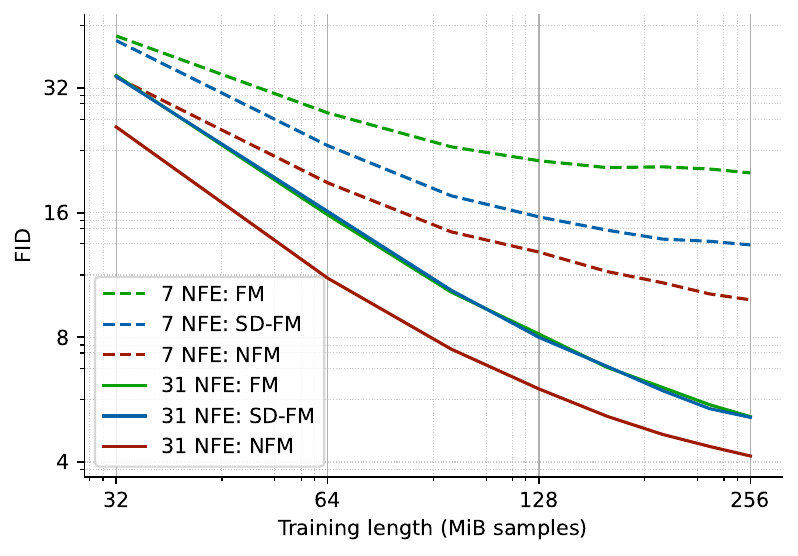}
        \caption{FID convergence on ImageNet64 without CFG}
        \label{plot:nocfg64}
    \end{center}
    \vskip -0.2in
\end{figure}

Thus far, all comparisons made use of guidance since it is what gives the best results and what is typically used in real world applications.
In this section, we take a look at unguided FID numbers.
Fig.\ref{plot:nocfg64} reports the FID evolution as training progresses for NFE=$31,7$.
As expected, while the absolute FID numbers are higher overall, the gains remain similar to the ones with CFG from Fig.\ref{plot:conv-fm64}.

\section{Discussions}
\label{sec:discuss}
\subsection{Optimal coupling}\label{sec:label-ot}

We consider multiple potential explanations as to why NF is outperforming SD-FM in the class-conditional setting:
\begin{itemize}[leftmargin=.2cm,itemsep=.0cm,topsep=0cm,parsep=2pt]
    \item \textbf{Label handling}: SD-FM handles labels by adding a fixed cost when the labels don't match between distributions $cost\cdot(y_1\neq y_2)$.
          This in itself is a workaround to fit the categorical label in the OT setting, but is not a pure OT handling because labels are not known topologically, only categorically.
          To illustrate this point, OT does not know that a cat label is closer to a dog than to a car.
          Meanwhile, NF uses the label to form a conditional bijection allowing it to learn from the data.
    \item \textbf{Inductive biases}: SD-FM only relies on learned potentials $g_i$ to produce assignment costs.
          Meanwhile, TarFlow performs a direct assignment using a learned neural network.
          Clearly TarFlow can produce more complex assignments.
          The student itself is also based on a neural network, albeit a different one from the teacher.
          It is possible, that the inductive biases of the teacher match better the inductive biases of the FM students than SD-FM.
          Following on that line of thinking, there could be other coupling methods yet to be discovered that would outperform both SD-FM and TarFlow by targeting the inductive biases of the students.
\end{itemize}

As previously mentioned, we see SD-FM and TarFlow as complementary.
While this is beyond the scope of this paper and left for further research, one can envision getting the best of both of worlds by combining them.
Consider that TarFlow maps real samples to a pseudo-Gaussian space which Gaussian-ness depends on the neural network capacity to minimize the objective (Eq.\ref{eq:nf_loss}) for the training data.
Therefore it might be advantageous to use SD-FM to map a true Gaussian space to TarFlow's pseudo-Gaussian one.
This would possibly preserve high quality generation by compensating for the less Gaussian spaces produced by NF models of lesser capacity or trained for shorter durations.

\subsection{Latent Encoding vs Noise Encoding}\label{sec:noise-encoding}

One could object that we did not account for the time to train the teacher and this is a valid concern: Training TarFlow is considerably more costly than solving a semi-discrete optimal transport problem as in SD-FM, and furthermore vanilla FM does not even need to learn a coupling in the first place before training.

If the goal is to produce a NF model used for purposes such as density/NLL estimation and to further distill this model to produce fast generation in addition then it is advantageous to use our method since the NF training cost is required regardless.

More broadly, NFs offer a way to encode data as noise analogous to how Auto-Encoders (AE) encode data as meaningful representations.
Indeed, while AEs are classically used to learn latent representations of data $x$, NFs do the same task for Gaussian representations, their bijective nature making them akin of lossless AEs.
This observation opens the door to offering pretrained foundation NF models for downstream applications such as data-noise coupling in the diffusion setting, just like is currently the case with AEs for latent representations.
Such pretrained NF models could be reused, just like pretrained AEs are currently reused for various tasks.
This is an exciting avenue for future work.

\section{Conclusion}
We presented \method{}, a new data-noise coupling method that accelerates training of FM models, while allowing fewer step sampling.
This method can also be seen as a distillation of NF models, in particular AR-NF models, not only drastically improving their sampling latency but also even improving the teacher's FID.
Our proposed method purposefully avoids the use of perceptual losses in order to remain general and potentially be applicable to other domains than images.
In addition, we provided an analysis of the $z$-space produced by NF, probing its inner properties from a neighborhood lens.
In this analysis, we showed that the NF pseudo-Gaussian space has a structure which does not preserve the input space neighborhood properties.

\textbf{Future work}:
While latency gains are very significant over AR-NF, further such gains are likely possible simply by distilling the learned FM models.
Along the same line of thought, application to Mean Flows \cite{geng2025mean} look like an interesting line of research, since their training might benefit more from lower noise.
More fundamentally, the structure of the $z$-space remains enigmatic due to the nearest neighbors in $x$-space not being the same in $z$-space.
One important question that needs to be elucidated is whether such structure is inherent to NF or not.
And a follow up question is whether or not preserving, even partially, the neighborhood property is really beneficial to FM.
As mentioned in Sec.\ref{sec:discuss}, combining SD-FM and \method{} also appears like an exciting prospect.
Looking ahead, \method{} could enable reusable foundation models that encode data as noise—much like AEs do for representations—and scale to text-image generation by leveraging teachers such as \cite{gu2025starflow}.

\section*{Impact Statement}

This paper presents work whose goal is to advance the field of Machine Learning.
There are many potential societal consequences of our work, none which we feel must be specifically highlighted here.

\section*{Acknowledgments}
We wish to thank Preetum Nakkiran, Huangjie Zheng, Barry Theobald and Miguel Angel Bautista Martin for their invaluable feedback.

\bibliography{main}
\bibliographystyle{icml2025}

\newpage
\appendix
\onecolumn
\section{Loss convergence curves.}

\begin{figure}[ht]
    \vskip 0.2in
    \centering
    \begin{subfigure}{0.49\textwidth}
        \includegraphics[width=\columnwidth]{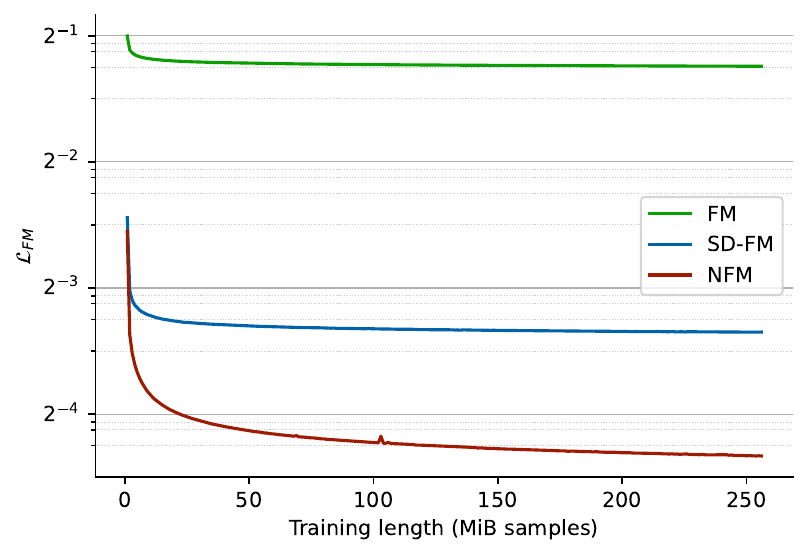}
        \caption{ImageNet64}
    \end{subfigure}
    \begin{subfigure}{0.49\textwidth}
        \includegraphics[width=\columnwidth]{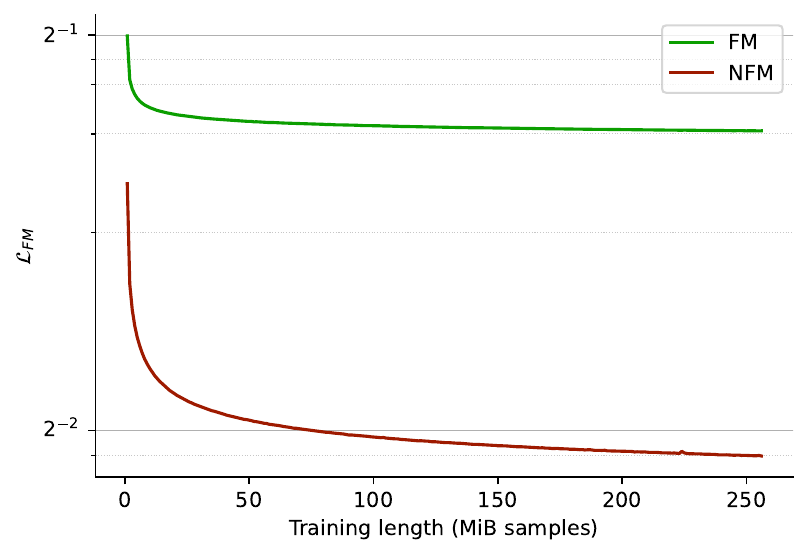}
        \caption{ImageNet256}
    \end{subfigure}
    \caption{FM loss convergence}
    \label{plot:conv-mse}
    \vskip -0.2in
\end{figure}

In Fig.\ref{plot:conv-mse}, we plot the loss convergence for FM, SD-FM and \method{}.
It is immediately clear that better couplings yield lower losses, so much so we resorted to plotting them in log-space to better visualize the magnitude differences.
While the loss gains are far greater than FID gains for $\text{NFE}=31$, these loss gains might explain why bigger FID gains are observed for lower NFE counts as shown in Tab.\ref{table:nfe-64}.

\section{Samples}
\clearpage
\begin{figure}[t]
    \centering
    \begin{subfigure}{1.\textwidth}
        \includegraphics[width=\textwidth]{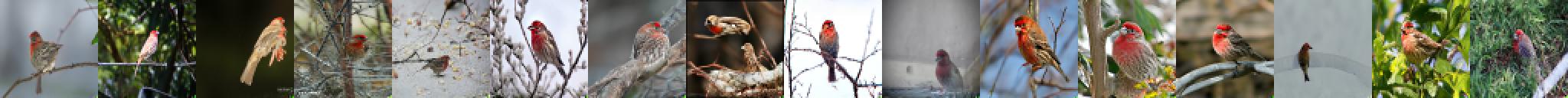}
        \caption{TarFlow Teacher}
    \end{subfigure}
    \\
    \begin{subfigure}{1.\textwidth}
        \includegraphics[width=\textwidth]{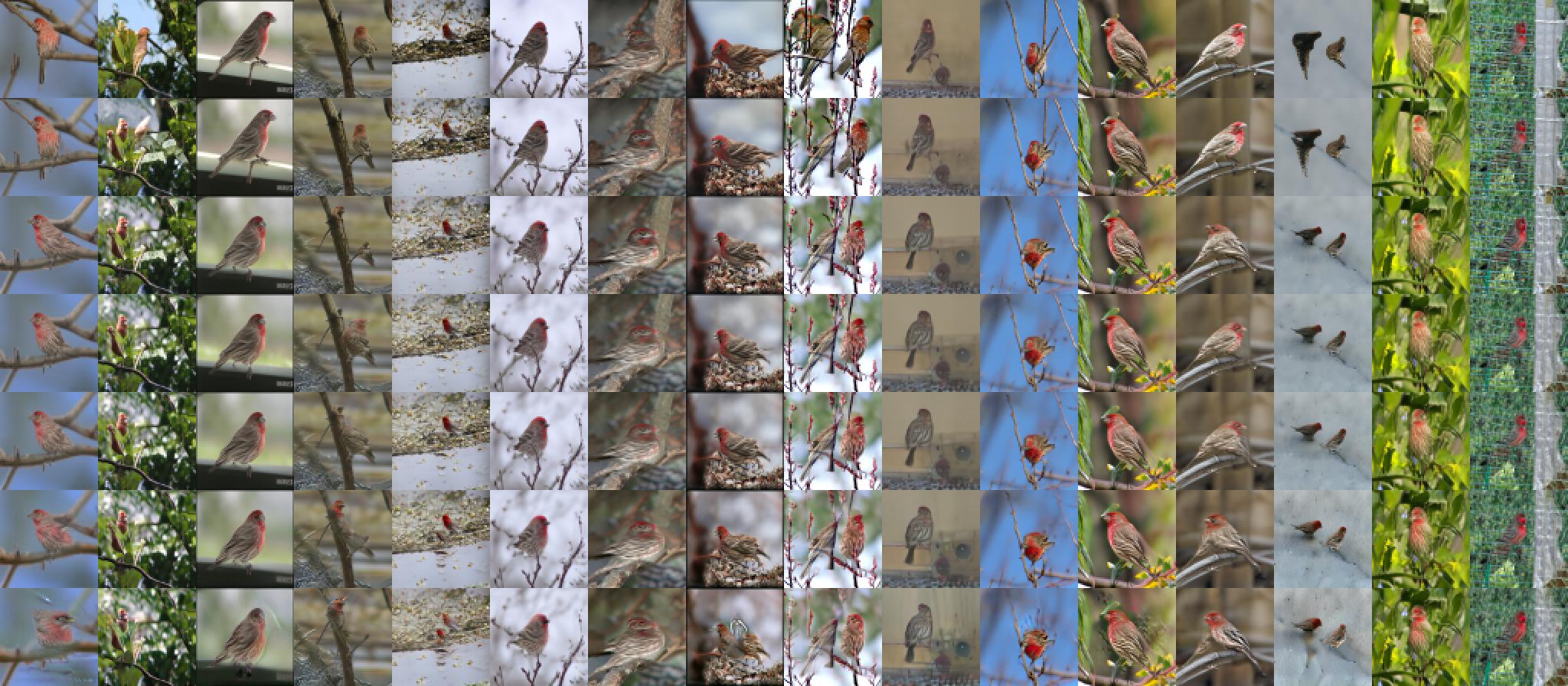}
        \caption{FM Student (rows $1,2,3$ show $31,15,7$ NFEs with Heun solver, rows $4,5,6,7$ show $6,5,4,3$ NFEs with Euler solver)}
    \end{subfigure}
    \vskip -0.1in
    \caption{ImageNet64 samples for class 12: house finch.}
    \label{image:64-house-finch}
    \vskip -0.2in
\end{figure}

\begin{figure}[t]
    \centering
    \begin{subfigure}{1.\textwidth}
        \includegraphics[width=\textwidth]{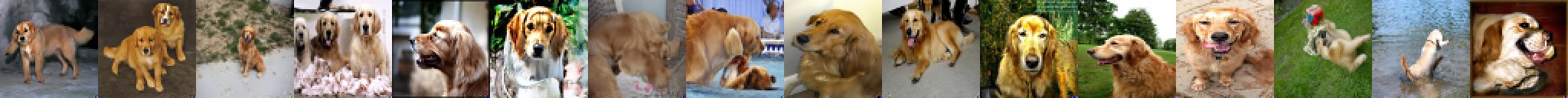}
        \caption{TarFlow Teacher}
    \end{subfigure}
    \\
    \begin{subfigure}{1.\textwidth}
        \includegraphics[width=\textwidth]{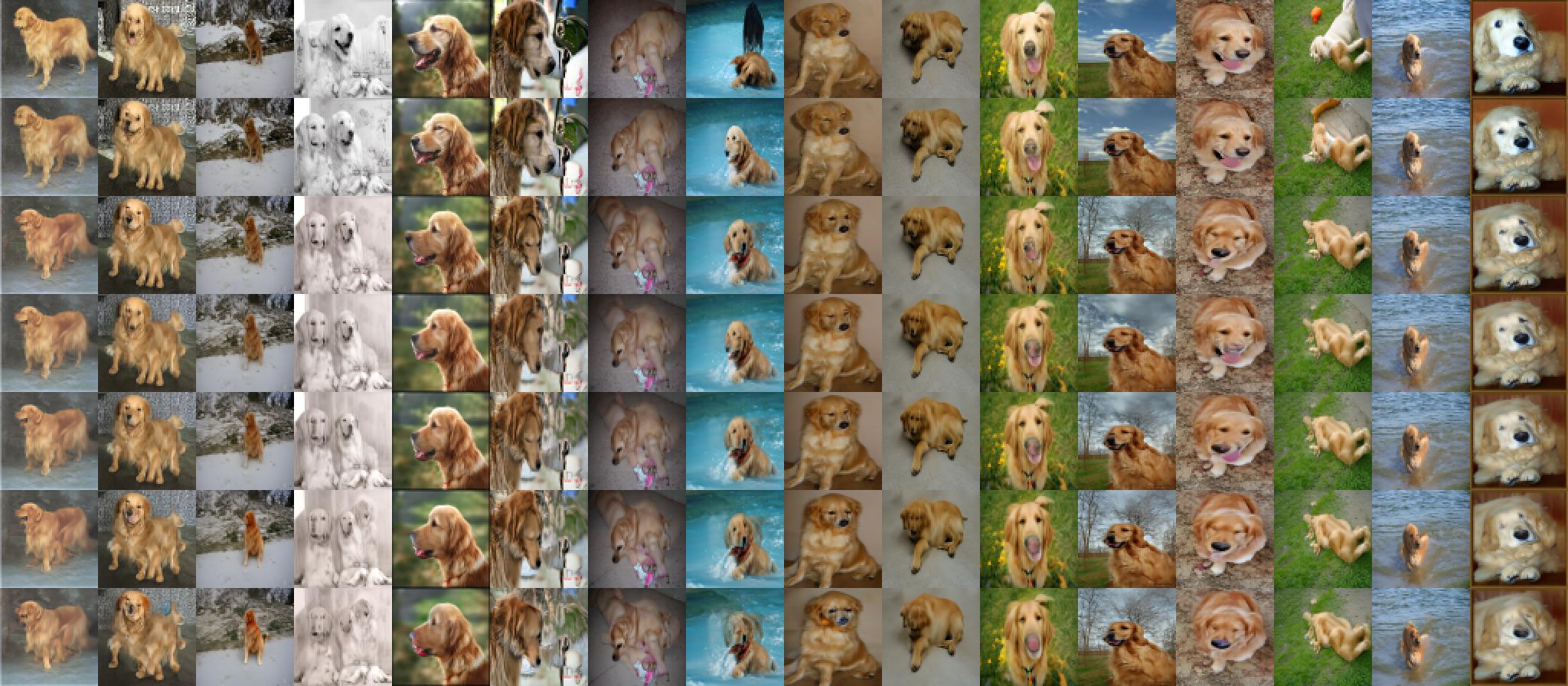}
        \caption{FM Student (rows $1,2,3$ show $31,15,7$ NFEs with Heun solver, rows $4,5,6,7$ show $6,5,4,3$ NFEs with Euler solver)}
    \end{subfigure}
    \vskip -0.1in
    \caption{ImageNet64 samples for class 207: golden retriever.}
    \label{image:64-golden-retriever}
    \vskip -0.2in
\end{figure}

\begin{figure}[t]
    \centering
    \begin{subfigure}{1.\textwidth}
        \includegraphics[width=\textwidth]{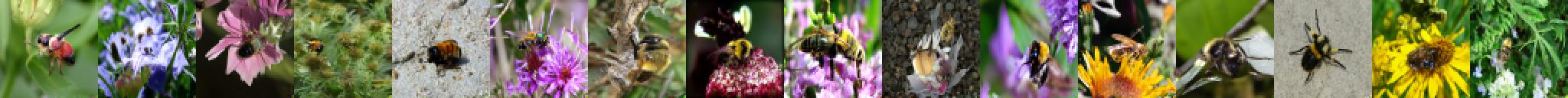}
        \caption{TarFlow Teacher}
    \end{subfigure}
    \\
    \begin{subfigure}{1.\textwidth}
        \includegraphics[width=\textwidth]{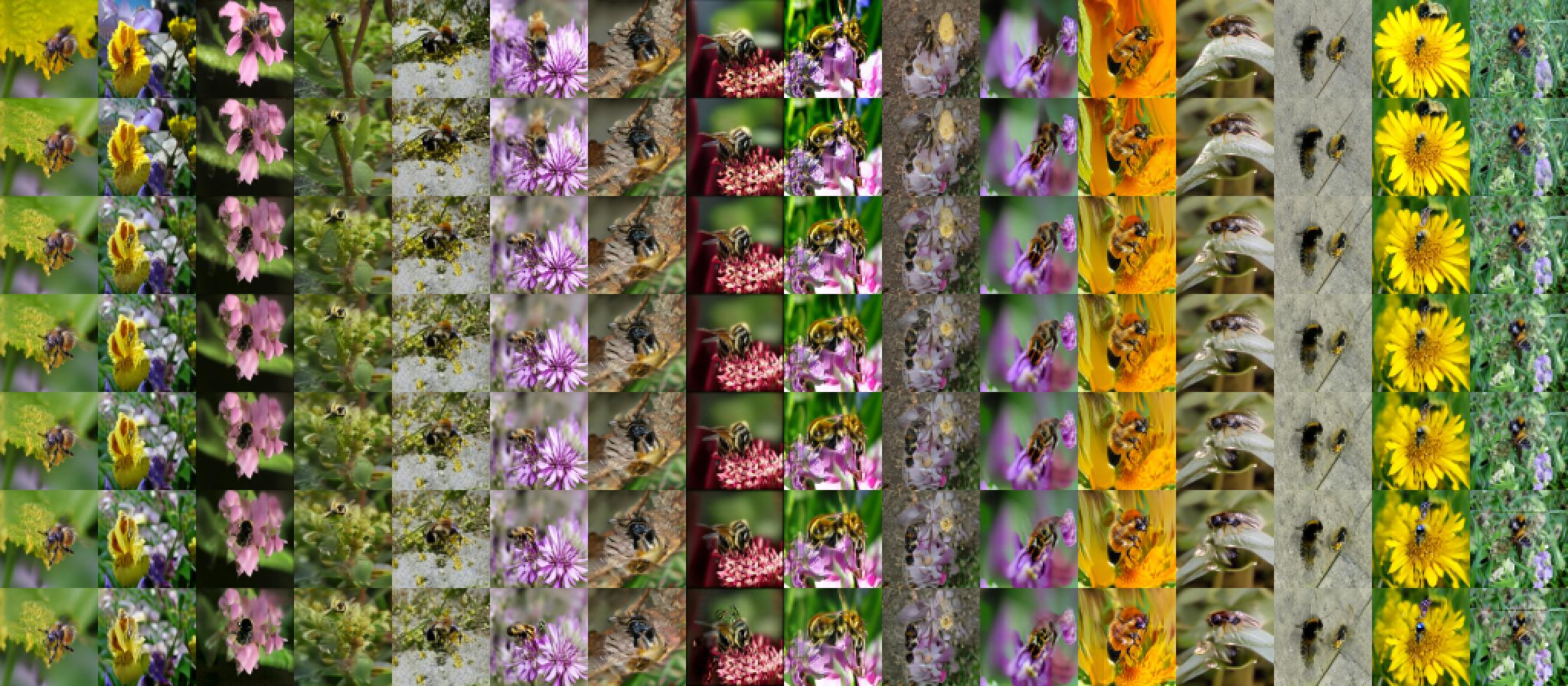}
        \caption{FM Student (rows $1,2,3$ show $31,15,7$ NFEs with Heun solver, rows $4,5,6,7$ show $6,5,4,3$ NFEs with Euler solver)}
    \end{subfigure}
    \vskip -0.1in
    \caption{ImageNet64 samples for class 309: bee.}
    \label{image:64-bee}
    \vskip -0.2in
\end{figure}

\begin{figure}[t]
    \centering
    \begin{subfigure}{1.\textwidth}
        \includegraphics[width=\textwidth]{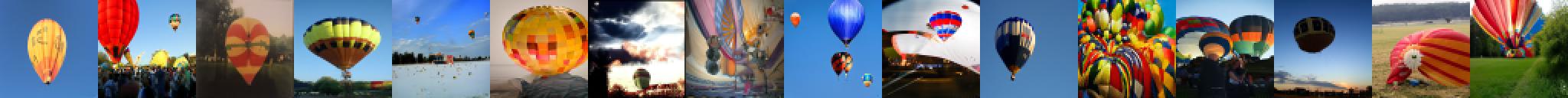}
        \caption{TarFlow Teacher}
    \end{subfigure}
    \\
    \begin{subfigure}{1.\textwidth}
        \includegraphics[width=\textwidth]{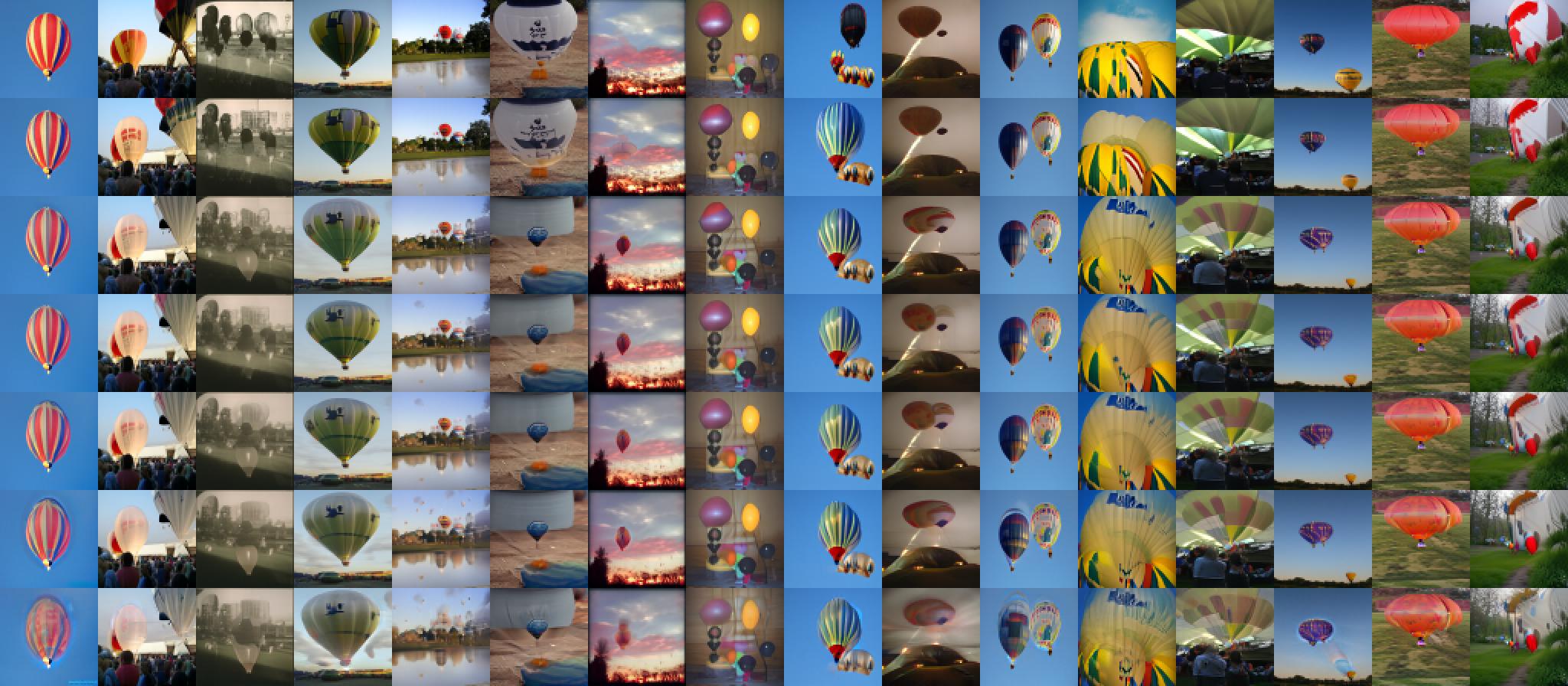}
        \caption{FM Student (rows $1,2,3$ show $31,15,7$ NFEs with Heun solver, rows $4,5,6,7$ show $6,5,4,3$ NFEs with Euler solver)}
    \end{subfigure}
    \vskip -0.1in
    \caption{ImageNet64 samples for class 417: balloon.}
    \label{image:64-balloon}
    \vskip -0.2in
\end{figure}

\begin{figure}[t]
    \centering
    \begin{subfigure}{1.\textwidth}
        \includegraphics[width=\textwidth]{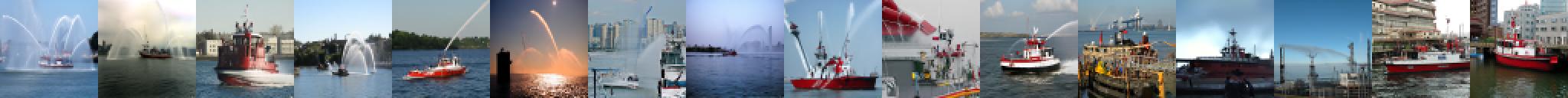}
        \caption{TarFlow Teacher}
    \end{subfigure}
    \\
    \begin{subfigure}{1.\textwidth}
        \includegraphics[width=\textwidth]{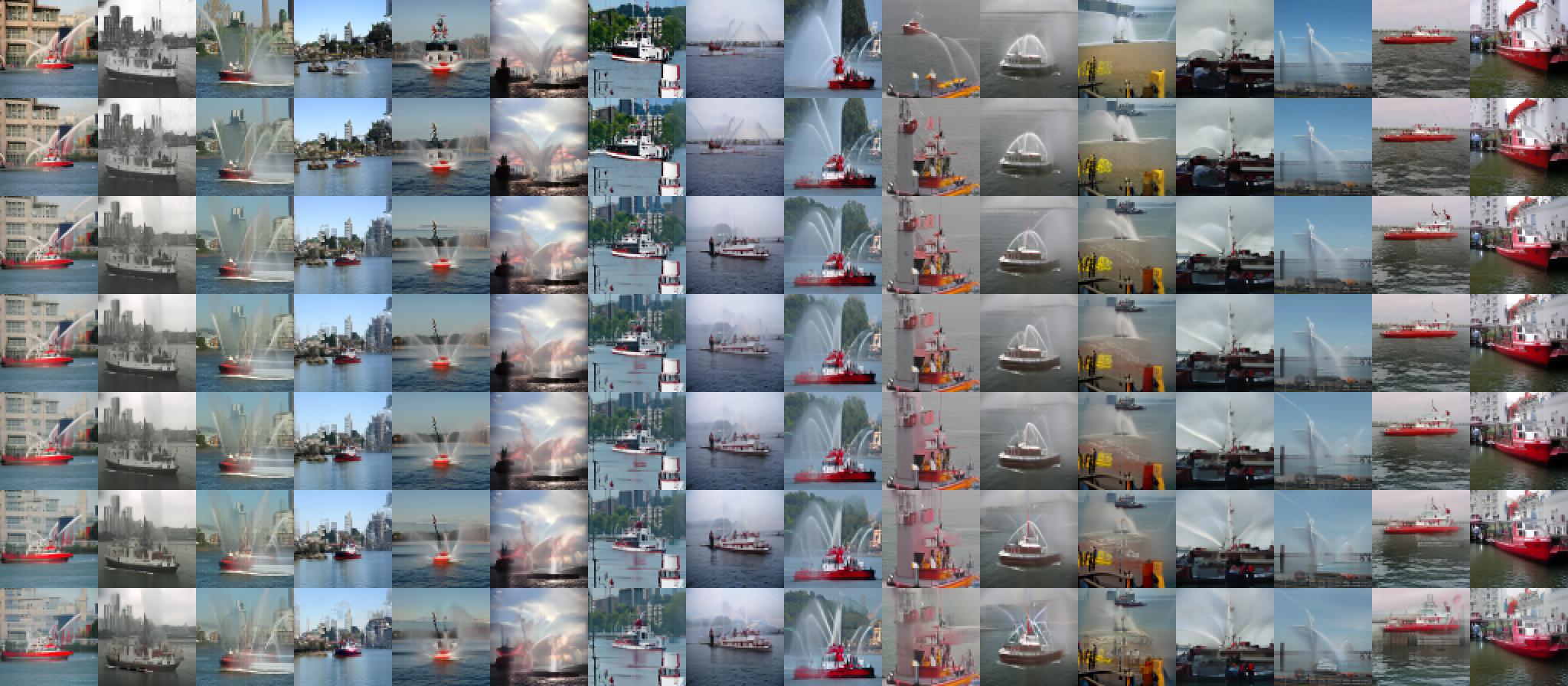}
        \caption{FM Student (rows $1,2,3$ show $31,15,7$ NFEs with Heun solver, rows $4,5,6,7$ show $6,5,4,3$ NFEs with Euler solver)}
    \end{subfigure}
    \vskip -0.1in
    \caption{ImageNet64 samples for class 554: fireboat.}
    \label{image:64-fireboat}
    \vskip -0.2in
\end{figure}

\begin{figure}[t]
    \centering
    \begin{subfigure}{1.\textwidth}
        \includegraphics[width=\textwidth]{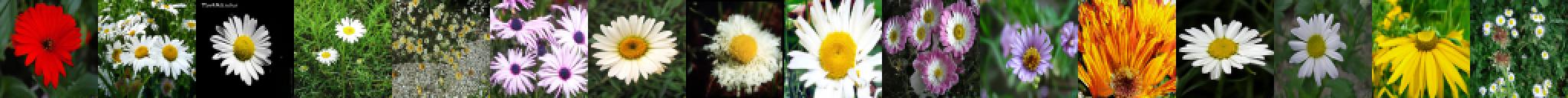}
        \caption{TarFlow Teacher}
    \end{subfigure}
    \\
    \begin{subfigure}{1.\textwidth}
        \includegraphics[width=\textwidth]{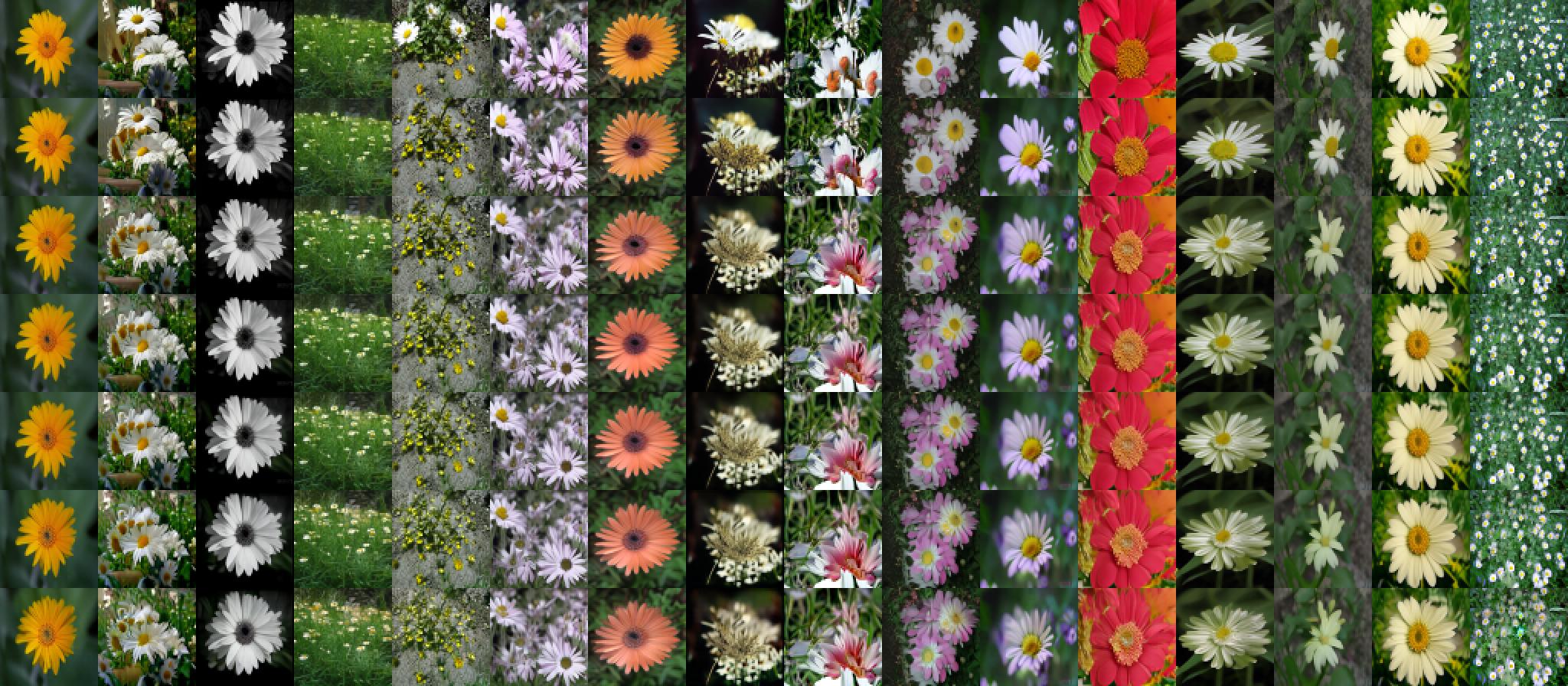}
        \caption{FM Student (rows $1,2,3$ show $31,15,7$ NFEs with Heun solver, rows $4,5,6,7$ show $6,5,4,3$ NFEs with Euler solver)}
    \end{subfigure}
    \vskip -0.1in
    \caption{ImageNet64 samples for class 985: daisy.}
    \label{image:64-daisy}
    \vskip -0.2in
\end{figure}

\clearpage
\begin{figure}[t]
    \vskip 0.2in
    \centering
    \begin{subfigure}{1.\textwidth}
        \includegraphics[width=\textwidth]{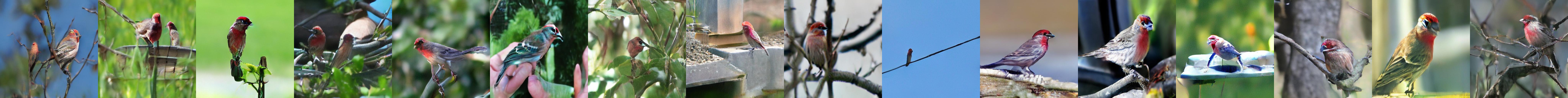}
        \caption{TarFlow Teacher}
    \end{subfigure}
    \\
    \begin{subfigure}{1.\textwidth}
        \includegraphics[width=\textwidth]{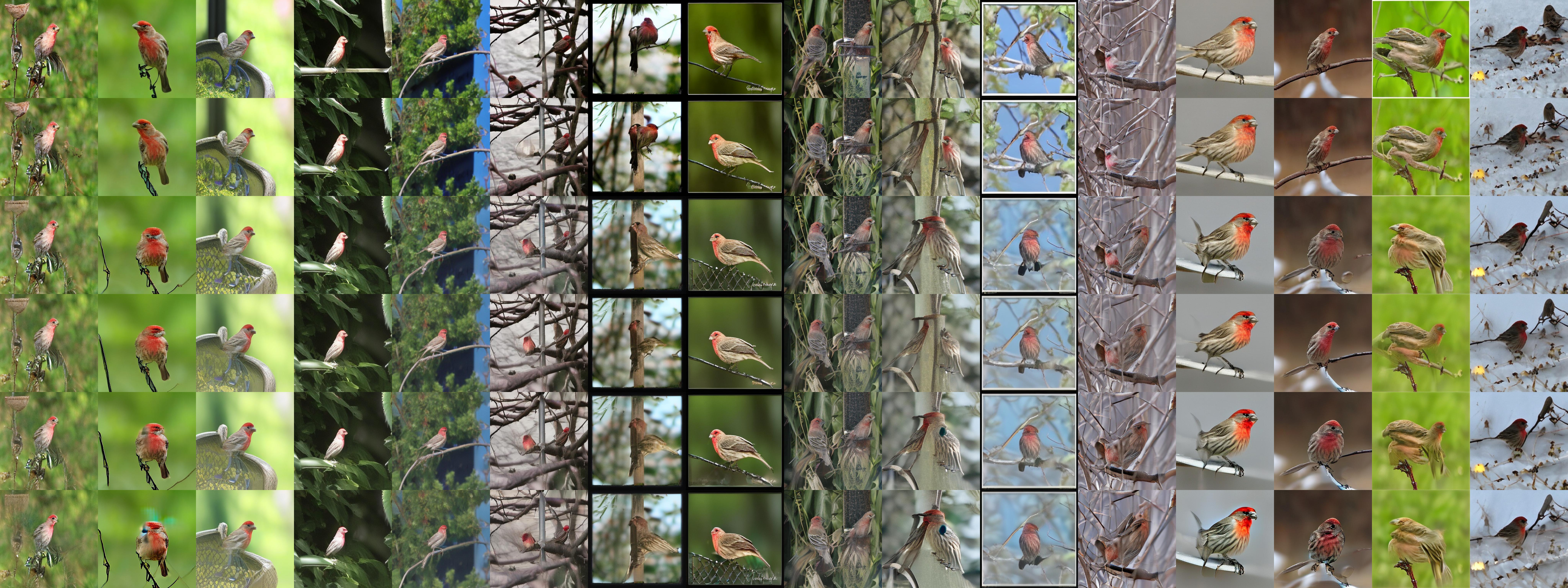}
        \caption{FM Student (rows $1,2,3$ show $31,15,7$ NFEs with Heun solver, rows $4,5,6$ show $6,5,4$ NFEs with Euler solver)}
    \end{subfigure}
    \vskip -0.1in
    \caption{ImageNet256 samples for class 12: house finch.}
    \label{image:256-house-finch}
    \vskip -0.2in
\end{figure}

\begin{figure}[t]
    \vskip 0.2in
    \centering
    \begin{subfigure}{1.\textwidth}
        \includegraphics[width=\textwidth]{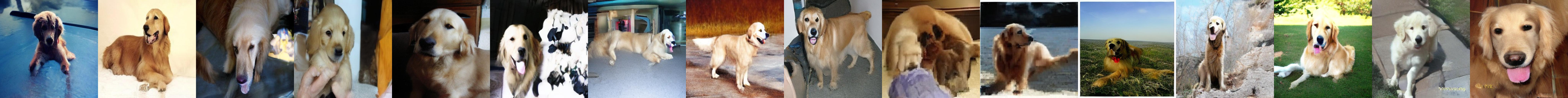}
        \caption{TarFlow Teacher}
    \end{subfigure}
    \\
    \begin{subfigure}{1.\textwidth}
        \includegraphics[width=\textwidth]{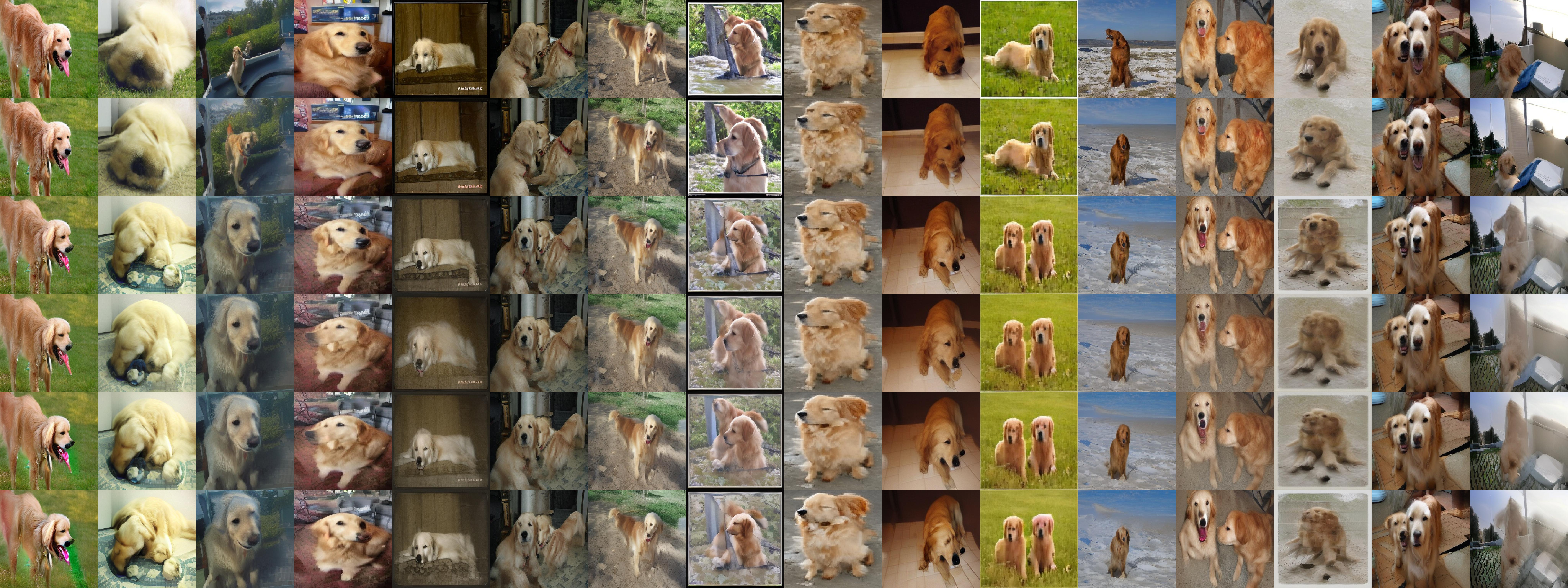}
        \caption{FM Student (rows $1,2,3$ show $31,15,7$ NFEs with Heun solver, rows $4,5,6$ show $6,5,4$ NFEs with Euler solver)}
    \end{subfigure}
    \vskip -0.1in
    \caption{ImageNet256 samples for class 207: golden retriever.}
    \label{image:256-golden-retriever}
    \vskip -0.2in
\end{figure}

\begin{figure}[t]
    \vskip 0.2in
    \centering
    \begin{subfigure}{1.\textwidth}
        \includegraphics[width=\textwidth]{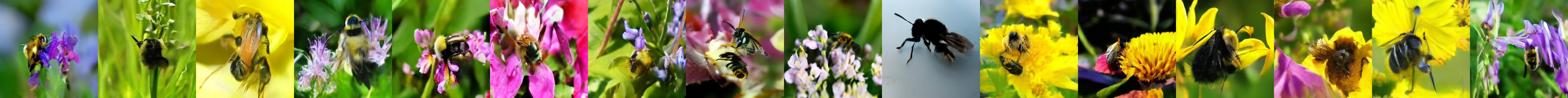}
        \caption{TarFlow Teacher}
    \end{subfigure}
    \\
    \begin{subfigure}{1.\textwidth}
        \includegraphics[width=\textwidth]{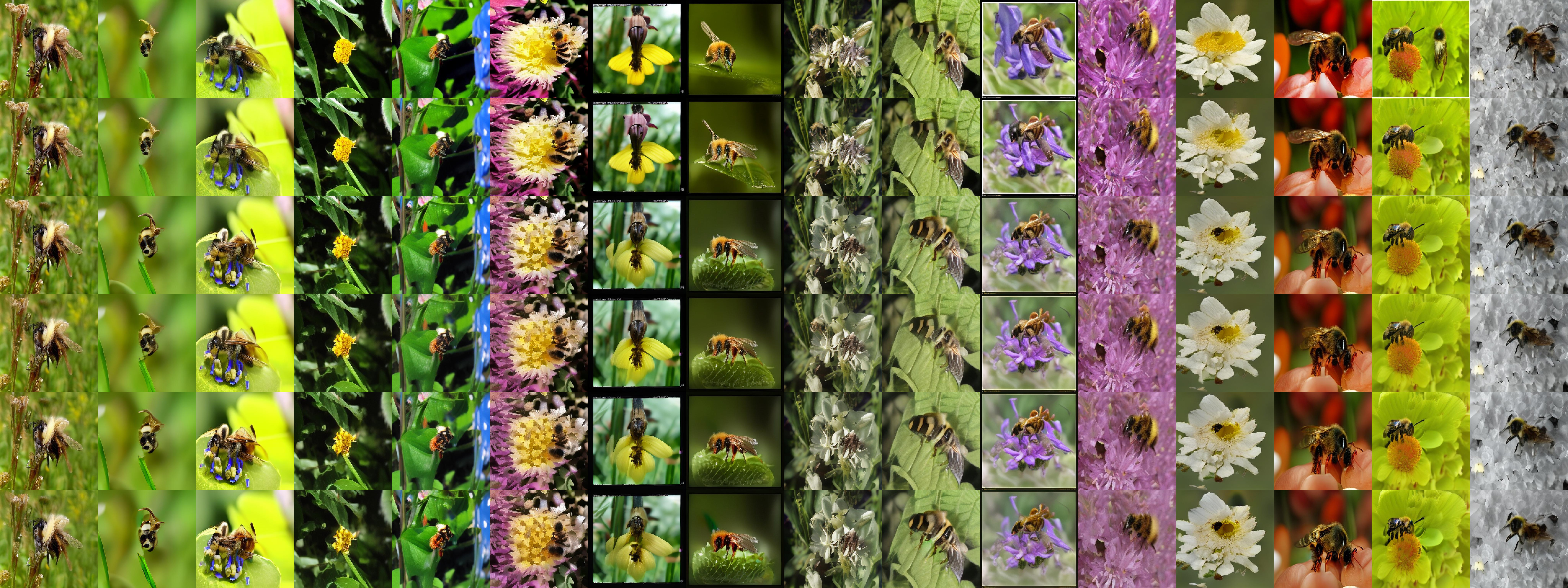}
        \caption{FM Student (rows $1,2,3$ show $31,15,7$ NFEs with Heun solver, rows $4,5,6$ show $6,5,4$ NFEs with Euler solver)}
    \end{subfigure}
    \vskip -0.1in
    \caption{ImageNet256 samples for class 309: bee.}
    \label{image:256-bee}
    \vskip -0.2in
\end{figure}

\begin{figure}[t]
    \vskip 0.2in
    \centering
    \begin{subfigure}{1.\textwidth}
        \includegraphics[width=\textwidth]{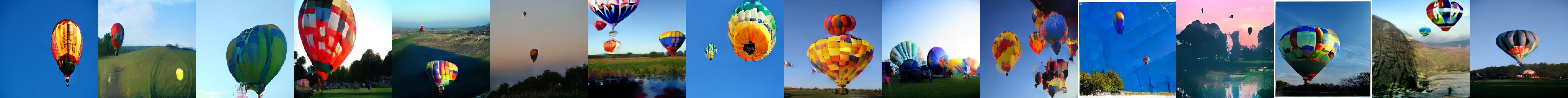}
        \caption{TarFlow Teacher}
    \end{subfigure}
    \\
    \begin{subfigure}{1.\textwidth}
        \includegraphics[width=\textwidth]{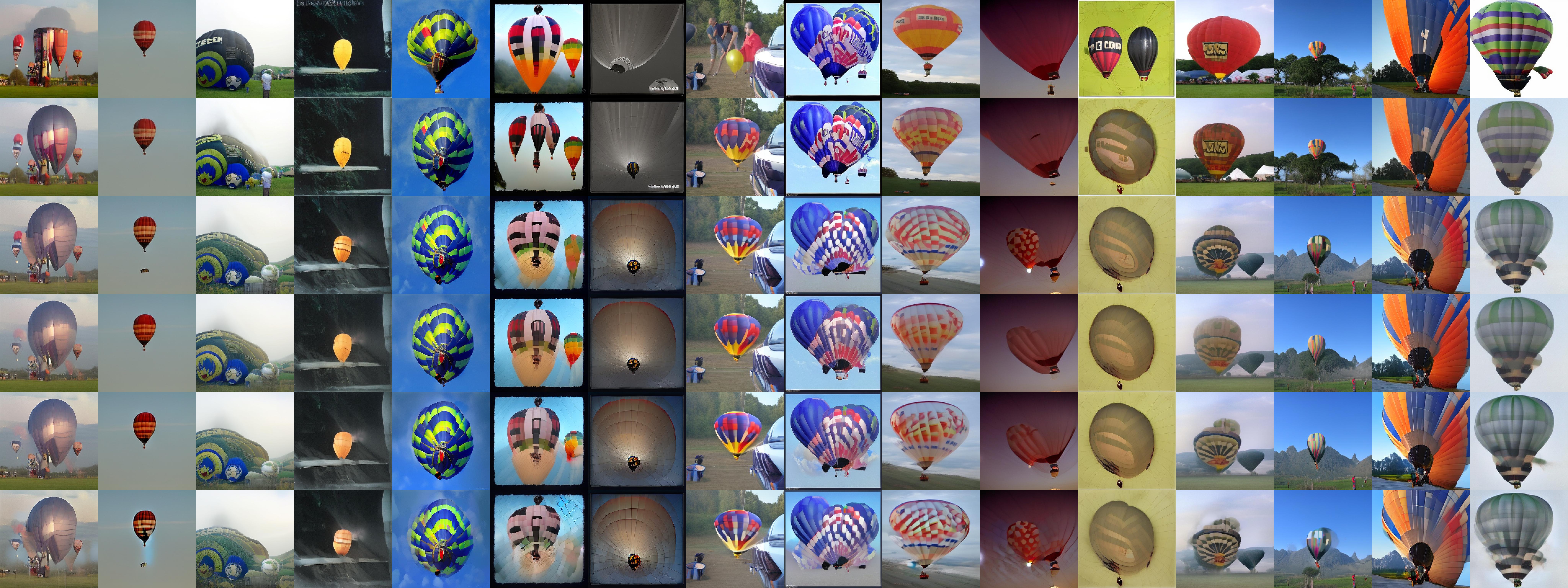}
        \caption{FM Student (rows $1,2,3$ show $31,15,7$ NFEs with Heun solver, rows $4,5,6$ show $6,5,4$ NFEs with Euler solver)}
    \end{subfigure}
    \vskip -0.1in
    \caption{ImageNet256 samples for class 417: balloon.}
    \label{image:256-balloon}
    \vskip -0.2in
\end{figure}

\begin{figure}[t]
    \vskip 0.2in
    \centering
    \begin{subfigure}{1.\textwidth}
        \includegraphics[width=\textwidth]{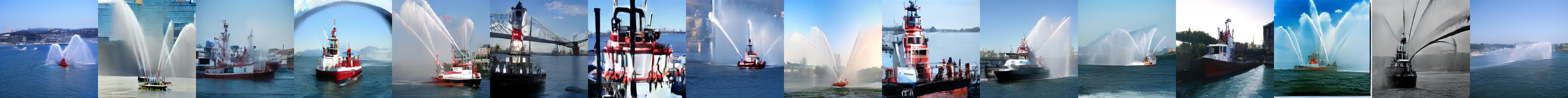}
        \caption{TarFlow Teacher}
    \end{subfigure}
    \\
    \begin{subfigure}{1.\textwidth}
        \includegraphics[width=\textwidth]{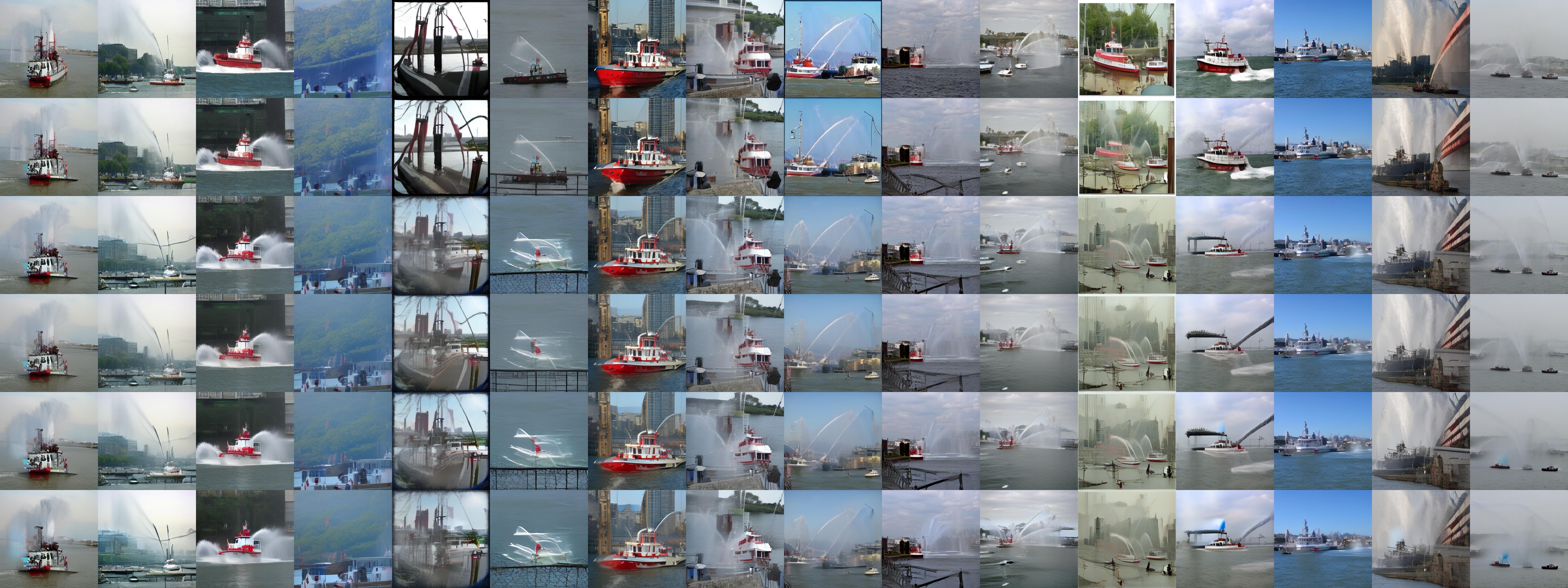}
        \caption{FM Student (rows $1,2,3$ show $31,15,7$ NFEs with Heun solver, rows $4,5,6$ show $6,5,4$ NFEs with Euler solver)}
    \end{subfigure}
    \vskip -0.1in
    \caption{ImageNet256 samples for class 554: fireboat.}
    \label{image:256-fireboat}
    \vskip -0.2in
\end{figure}

\begin{figure}[t]
    \vskip 0.2in
    \centering
    \begin{subfigure}{1.\textwidth}
        \includegraphics[width=\textwidth]{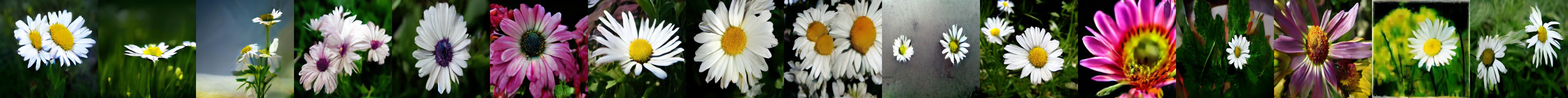}
        \caption{TarFlow Teacher}
    \end{subfigure}
    \\
    \begin{subfigure}{1.\textwidth}
        \includegraphics[width=\textwidth]{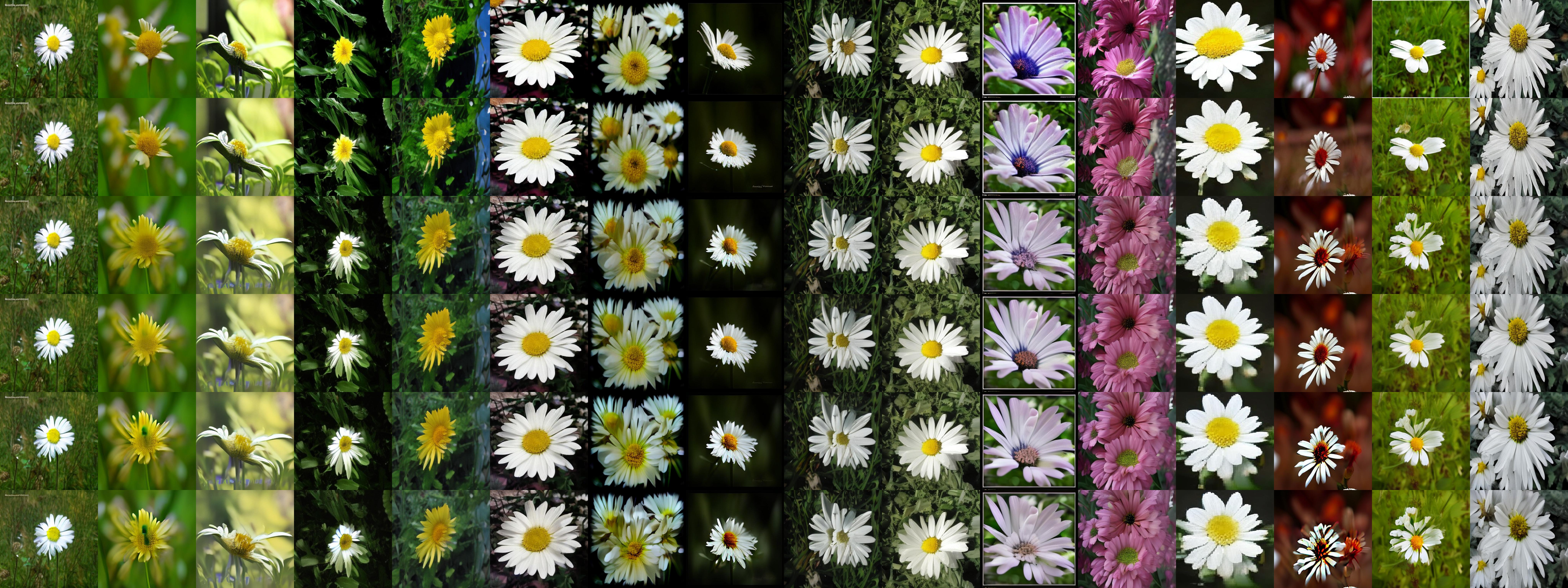}
        \caption{FM Student (rows $1,2,3$ show $31,15,7$ NFEs with Heun solver, rows $4,5,6$ show $6,5,4$ NFEs with Euler solver)}
    \end{subfigure}
    \vskip -0.1in
    \caption{ImageNet256 samples for class 985: daisy.}
    \label{image:256-daisy}
    \vskip -0.2in
\end{figure}

\end{document}